\documentclass{article}

\usepackage{arxiv}

\usepackage[utf8]{inputenc} 
\usepackage[T1]{fontenc}    
\usepackage{hyperref}       
\usepackage{url}            
\usepackage{booktabs}       
\usepackage{amsfonts}       
\usepackage{nicefrac}       
\usepackage{microtype}      
\usepackage{lipsum}		
\usepackage{graphicx}
\usepackage{natbib}
\usepackage{doi}
\usepackage{xcolor}
\usepackage{caption}
\usepackage{subcaption}

\title{{\tt MigrationsKB}: A Knowledge Base of Public Attitudes towards Migrations and their Driving Factors}

\author{ {\hspace{1mm}Yiyi Chen}, {\hspace{1mm}Harald Sack}, {\hspace{1mm}Mehwish Alam} \\
\texttt{firstname.lastname@fiz-karlsruhe.de} \\
FIZ Karlsruhe -- Leibniz Institute for Information Infrastructure, Germany \\
Karlsruhe Institute of Technology, Institute AIFB, Germany
}

\begin{document}
\maketitle

\begin{abstract}
	
With the increasing trend in the topic of migration in Europe, the public is now more engaged in expressing their opinions through various platforms such as Twitter. Understanding the online discourses is therefore essential to capture the public opinion. The goal of this study is the analysis of social media platform to quantify public attitudes towards migrations and the identification of different factors causing these attitudes. The tweets spanning from 2013 to Jul-2021 in the European countries which are hosts to immigrants are collected, pre-processed, and filtered using advanced topic modeling technique. BERT-based entity linking and sentiment analysis, and attention-based hate speech detection are performed to annotate the curated tweets. Moreover, the external databases are used to identify the potential social and economic factors causing negative attitudes of the people about migration. To further promote research in the interdisciplinary fields of social science and computer science, the outcomes are integrated into a Knowledge Base (KB), i.e., {\tt MigrationsKB} which significantly extends the existing models to take into account the public attitudes towards migrations and the economic indicators. This KB is made public using FAIR principles, which can be queried through SPARQL endpoint\footnote{\url{https://mgkb.fiz-karlsruhe.de/sparql/}}. Data dumps are made available on Zenodo\footnote{\url{https://bit.ly/2W01Doc}}. 
\end{abstract}


\section{Introduction}
\label{sec:introduction}

Social media has become one of the most widely used and important channel for people to express their opinions about the events happening around the globe. It is one of the most useful sources for measuring the attitudes of the public on important topics such as migration, climate change, green deal, etc. More specifically, migration has become one of the mainstream topics in Europe. Many European Projects have recently been focusing on the similar topic from different perspectives such as PERCEPTIONS\footnote{\url{https://project.perceptions.eu/}}, METICOS\footnote{\url{https://meticos-project.eu/}}, etc.
Many efforts have been put into studying the attitudes of public towards migrations from various perspectives~\citep{hainmueller2014, lenka2018, dempster2020}. This study in particular focuses on analyzing social media platform in order to quantify and study public attitudes towards migrations and identify different factors which could be probable causes of these attitudes. This study utilizes advanced Artificial Intelligence (AI) methods based on knowledge graphs and neural networks such as contextual language models for analyzing these public attitudes on social media platform such as Twitter. This study aims at:
(a) providing better understanding of public attitudes towards migrations, (b) explain possible reasons why these attitudes towards migrations are what they are, (c) define a Knowledge Base (KB) called as {\tt MigrationsKB} built by taking into account the semantics underlying this field of study, (d) publish this resource using FAIR principles~\citep{fair}, i.e., make the resource Findable, Accessible, Interoperable, and Reusable.

Since the study mainly focuses on the analysis of social media platform such as Twitter, many kinds of challenges arise, i.e., millions of tweets in noisy natural language are being posted around the globe about a particular topic each day, which makes it impossible for the humans to process this information, leading to the necessity of automated processing. Many efforts have been conducted for creating KB for integrating twitter data, such as, TweetsKB~\citep{tweetskb}, which is a KB of Twitter data in general, TweetsCOV19~\citep{tweetscov19}, which uses COVID-19 related tweets from TweetsKB. One of the other analytical tool which is related to the topic of migration is MigrAnalytics~\citep{AlamGeseseRezaie2020}, which analyzes tweets about migrations from TweetsKB.

Instead of using a subset of TweetsKB, {\tt MigrationsKB} extends it by focusing on the specific aspects regarding the topic of migration and providing advanced deep learning based semantic annotations. It will help in facilitating further research in the field of social sciences, and to provide a viable corpus for further analysis in the domain of migration.


In order to analyze the public attitudes towards migrations in the destination countries in Europe, the geotagged tweets are extracted using migration related keywords. The irrelevant tweets are then filtered by using state-of-the-art neural network based topic modeling. It further utilizes contextualized word embeddings~\citep{corr/abs-2003-07278} and transfer learning for sentiment analysis and hate speech detection. Temporal and geographical dimensions are then explored for measuring the public attitudes towards migrations in a certain period of time in a certain region. Entity linking is applied to identify the entity mentions linked to Wikipedia for enabling easy search over the tweets related to a particular topic. In order to identify the potential social and economic factors driving the migration flows, external databases such as Eurostat\footnote{\url{https://ec.europa.eu/eurostat}} and Statista\footnote{\url{https://www.statista.com/}},  are used to analyze the correlation between the public attitudes and the established economic indicators in a specific region in a certain time-period.

In order to enable the reusability of the results of this analysis, the outcome is then integrated into {\tt MigrationsKB} which is an extension of the RDFS\footnote{\url{https://www.w3.org/TR/rdf-schema/}} model as originally defined in TweetsKB. It is extended by defining new classes and entities to cover the Geo information of the tweets, the results of hate speech detection as well as to integrate the information about the Economic Indicators which could be the potential cause of negativity or hatred towards migrations. Finally, 
the competency questions are defined, the answers to which can be retrieved with the help of SPARQL\footnote{\url{https://www.w3.org/TR/rdf-sparql-query/}} queries. The source code has been made public for reproducibility reasons and is available through a GitHub repository\footnote{\url{https://github.com/MigrationsKB/MGKB}}. Information related to {\tt MigrationsKB} is available through the web page\footnote{\url{https://MigrationsKB.github.io/MGKB/}}. The SPARQL endpoint of {\tt MigrationsKB} is also made publicly available to enable querying. The dump of annotated data is available through Zenodo.

This paper is structured as follows: In Section~\ref{sec:related_work} discusses the related work. In  Section~\ref{sec:resource_description}, the details of the resource are presented. Section~\ref{sec:MigrationsKB} provides details of {\tt MigrationsKB}, the relevant competency questions are discussed. Finally, section~\ref{sec:discussion} concludes the paper and discusses the future work.

\section{Related Work}
\label{sec:related_work}


This section discusses studies which combine KBs and Twitter information from various domains in the first part. The second part of this section discusses the studies conducted for assessing the public attitudes towards migrations.  The third part discusses the European projects involved in the topic of migration. 

\subsection{Knowledge Bases based on Twitter Data} Several studies have been conducted which provide a KB containing Tweets from a particular time span for making it more usable by researchers. TweetsKB~\citep{tweetskb} is one such KB which contains more than 1.5 billion tweets spanning almost 5 years, including entity and sentiment annotations, and provides a publicly available RDF dataset using established vocabularies to further facilitate different data exploration scenarios, such as entity-centric sentiment analysis and temporal entity analysis, etc. In the event of COVID-19 pandemic, TweetsCOV19~\citep{tweetscov19}, deploying the RDF schema of TweetsKB, provides a knowledge base of COVID-19-related tweets, building on a TweetsKB subset spanning from October 2019 to April 2020. The study applies the same feature extraction and data publishing methods as TweetsKB. 

As a step forward in combining KB and Twitter information to the field of analyzing migration related data, MigrAnalytics~\citep{AlamGeseseRezaie2020} is introduced. It uses TweetsKB as a starting point to select data during the peak migration period from 2016 to 2017. MigrAnalytics analyzes tweets about migrations from TweetsKB including the hashtags and entities from the single seed word ``Refugee" and then further combines European migration statistics to correlate with the selected tweets. MigrAnalytics enriches the keywords using WordNet, Wikipedia, and Word Embeddings. However, it uses a very naive algorithm for performing sentiment analysis. Moreover, it does not introduce any sophisticated way to remove the irrelevant tweets. In contrast, the methods used for generating {\tt MigrationsKB} follow more advanced methods based on neural networks for sentiment analysis as well as hate speech detection and extends the RDFS model with relevant information. Also, due to the low Recall (i.e.,39\%) of the entity linking~\citep{tweetscov19}, the entities extracted in TweetsKB can not guarantee a comprehensive curation of migration related tweets. For the current study, three rounds of crawling are conducted (cf. Section~\ref{sec:curation}).


\subsection{Public Attitudes Towards Migrations}
While the prominence of the topic of migration has risen sharply over the last decade in Europe, many efforts have been invested into analyzing the public attitudes towards migrations from various aspects. For instance, \citep{hainmueller2014} is based on the studies conducted during the last 2 decades explaining public attitudes on immigration policy in North America and Western Europe. The authors investigate the natives' attitudes towards immigration from perspectives of political economy and political psychology.

It is found that, attitudes towards immigration are shaped by sociotropic concerns about its cultural impacts - and to a lesser extent its economic impacts - on the nation as a whole. While in \citep{lenka2018}, the authors explore the academic literature and the most up-to-date data across 17 countries on both sides of the Mediterranean. The study summarizes theoretical explanations for attitudes towards immigration including media effects, economic competition, contact and group threat theories, early life socialization effects, and psychological effects. It also concludes that in Europe, attitudes towards immigration are notably stable, rather than becoming more negative. More recently, \citep{dempster2020} emphasizes on the factors of individuals' values and worldview. It states that individual factors (i.e., personality, early life norm acquisition, tertiary education, familial lifestyle and personal worldview) have more stable and strong impact on the person's attitudes towards immigration rather than the influence from politicians and media.

In \citep{lenka2018} and \citep{dempster2020}, the survey data is used exclusively, while for \citep{hainmueller2014} a comprehensive assessment of approximately 100 studies, including both survey and field experiment data, is conducted. However, for the current work, the analysis is performed on data from social media with automated approaches.

\subsection{Projects on Migrations in Europe}  Since migration has become one of the most popular and controversial topics in Europe, many projects have been established to gain perspectives and aid policy decisions regarding migrations. The PERCEPTIONS project aims to identify narratives, images, and perceptions of Europe abroad and to investigate how the discrepancies in different narratives lead to unrealistic expectations, problems, and security threats for both host countries and migrants. Eventually, it provides toolkits using all the above-mentioned information and measures to counteract the social issues. Meanwhile, METICOS project is mainly focused on creating a holistic solution to solve challenges for border management in the European Union.

\section{Knowledge Base Construction of Migration Related Tweets}
\label{sec:resource_description}

\begin{figure*}
    \centering
    \caption{Overall Framework for Constructing {\tt MigrationsKB}.}
    \includegraphics[width=0.8\textwidth]{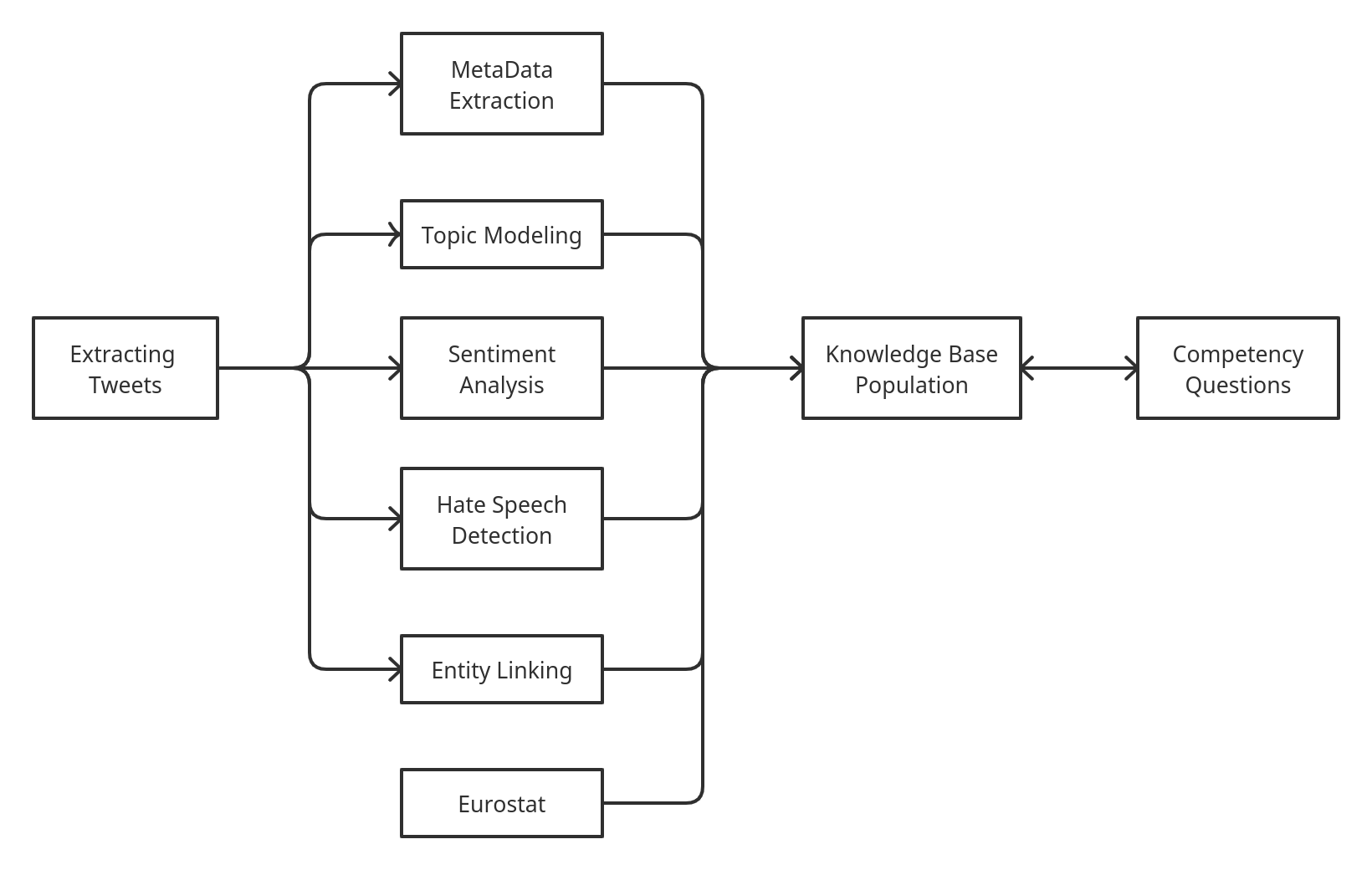}
    \label{fig:pipeline}
\end{figure*}

{\tt MigrationsKB} is an extension over TweetsKB with the specific focus on the topic of migration (as the name depicts). The goal of this KB is two-fold: (i) to provide a semantically annotated, query-able resource about public attitudes on social media towards migrations, (ii) to provide an insight into which factors in terms of economic indicators are the cause of that attitude. In order to achieve these goals, an overall framework for constructing is shown in Figure~\ref{fig:pipeline}. The first step is to define migration related keywords and perform keyword based extraction of geotagged tweets. The meta-data of the tweets is then extracted. Furthermore, topic modelling is performed for refining the tweets in case if irrelevant tweets are crawled in the tweet extraction phase. Contextual Embeddings are then used for performing sentiment analysis (i.e., tweets are classified into positive, negative and neutral) on the relevant set of tweets obtained after topic modeling. In order to further analyze the negative sentiments in terms of hate speech against the immigrants/refugees, tweets are further classified into three classes, i.e., hate, offensive, and normal. In order to achieve the second goal of this study, an analysis of factors causing the negative sentiment or the hatred against immigrants/refugees is performed with the help of plots. 

In order to make this information query-able with the help of SPARQL queries,  {\tt MigrationsKB}, a KB containing public attitudes towards migrations along with the factors driving these attitudes, is constructed.  {\tt MigrationsKB} is then populated with information extracted using the previously described framework. The statistics about these relevant factors such as unemployment rate, the gross domestic product growth rate (GDPR), etc. are extracted from Eurostat, Statista, UK Parliament\footnote{\url{https://www.parliament.uk/}}, and Office for National Statistics\footnote{\url{https://www.ons.gov.uk/}}.

\subsection{Collecting Migration Related Tweets}
\label{sec:curation}


In order to identify the public attitudes towards migrations in the EU countries, the first step is to select a list of destination countries, i.e., the countries hosting the immigrants/refugees. In order to do so, the statistics about asylum applications (annual aggregated) present on Eurostat \footnote{\url{https://ec.europa.eu/eurostat/databrowser/view/tps00191/default/table?lang=en}} are used to obtain the countries with higher frequency of asylum applications during the period from 2013 to 2020, as shown in Table~\ref{tab:statistics_asylum_seeker}. The list of countries includes: Germany, Spain, Poland, France, Sweden, United Kingdom, Austria, Hungary, Switzerland, Netherlands, and Italy. 

\begin{table*}[]
\centering
\caption{Statistics of the EU countries with the most first time asylum applicants.}
\resizebox{0.8\textwidth}{!}{\begin{tabular}{|c|c|c|c|c|c|c|c|c|c|}
\hline 
& \textbf{2013} & \textbf{2014}  & \textbf{2015}  &	\textbf{2016}  &	\textbf{2017}  &	\textbf{2018}  &	\textbf{2019}  &	\textbf{2020}  &	\textbf{SUM} \\\hline 
Germany &	126705 &	202645 &	476510 & 745160 &	222565 &	184180 &	165615 &	121955 &	2245335 \\\hline 
Spain	& 4485 &	5615 &	14780 &	15755 &	36610 &	54050 &	117800 &	88530 &	337625 \\\hline
Poland & 	15240 & 	8020 &	12190 &		12305 &		5045 &		4110 &		4070 &		2785 &		63765 \\\hline
France 	& 66265	 & 64310 &	76165 &	84270 &	99330 &	137665 &	151070 &	93470 &	772545  \\\hline 
Sweden	 & 54270  &	81185  &	162450 & 28795  &	26330 & 21560  & 26255  & 16225  &	417070 \\\hline 
United Kingdom  & 	30585  & 	32785  & 	40160  & 	39735  & 	34780	 &  38840 & 	46055 & 	36041 & 	298981\\\hline 
Austria & 	17500 & 	28035 & 	88160 & 	42255 & 	24715 & 	13710 & 	12860 & 	14180 & 241415 \\\hline 
Hungary	 & 18895  &	42775  & 177135  &	29430  &	3390  &	670	 & 500  &	115	 & 272910 \\\hline 
Switzerland	& 21305	 & 23560 &	39445 &	27140 &	18015 &	15160 &	14195 &	10990 &	169810 \\\hline
Netherlands	 & 13065 &	24495 &	44970 &	20945 &	18210 &	24025 &	25200 &	15255 &	186165  \\\hline

Italy &	26620 &	64625 &	83540 &	122960 &	128850 &	59950 &	43770 &	26535 &	556850 \\\hline 
\end{tabular}}
\label{tab:statistics_asylum_seeker}
\end{table*}


In the second step, relevant tweets are extracted using keywords related to the topic of immigration and refugees 
using word embeddings. The words ``immigration" and ``refugee" are used as the seed words based on which top-50 most similar words are extracted using pretrained Word2Vec model on Google News\footnote{\url{https://github.com/mmihaltz/word2vec-GoogleNews-vectors}} 
as well as fastText embeddings\footnote{\url{https://fasttext.cc/docs/en/crawl-vectors.html}}. These keywords are then manually filtered for relevance. Based on these keywords, a first round of crawling the tweets is performed. Then for the second round, most popular hashtags, i.e., the hashtags occurring in more than 100 crawled tweets are selected. These hashtags are also verified manually for relevance and then are used for crawling tweets in the second round. For both rounds, the crawled tweets span from 2013 to 2020. The third round of crawling is conducted, using both keywords and hashtags, to extract the tweets spanning from January to the end of July in 2021. The set of keywords, selected popular hashtags for filtering tweets are available on the GitHub repository\footnote{Keywords:\url{https://bit.ly/3APiKIw}, Hashtags: \url{https://bit.ly/2W1TMXk}}. The 10 most frequently occurring hashtags containing ``refugee" and ``immigrant" are shown in Table~\ref{tab:hashtag_refugee}.




The extracted tweets are further filtered using their geographical information,
i.e., only those tweets are selected which are geotagged with previously identified destination countries. 
About 66\% of the crawled tweets have exact coordinates in their Geo metadata, the rest contain place names, such as ``Budapest, Hungary". 

\begin{table*}[]
    \centering
    \caption{Statistics of Tweets.}
    \resizebox{\textwidth}{!}{
    \begin{tabular}{|c|c|c|c|c|c|c|c|c|c|c|c|c|}
    \hline 
                    & \textbf{Germany} & \textbf{Spain} & \textbf{Poland} & \textbf{France} & \textbf{Sweden} & \textbf{UK} & \textbf{Austria} & \textbf{Hungary} & \textbf{Switzerland} & 
                    \textbf{Netherlands} & \textbf{Italy} & \textbf{SUM} \\\hline
    1st round crawling  &  8,209 &  5,902  & 3,069 & 7,847   &   2,790 & 167,240 &  1,055 &  623 &  4,272 & 3,587 & 5,402   & 209,996 \\\hline 
    2nd round crawling &  17,031 & 14,981 & 2,986 & 20,202 & 4,116   & 78,074  & 5,063  & 2,768 & 8,169 &  11,943 & 23,718  &  189,051 \\\hline 
    3rd round crawling & 2,551 & 980 & 219 & 1,742 & 951 & 29,065 & 479 & 81 & 663 & 1,079 & 1,752 & 39,562 \\\hline 
    All (Unique)     & 26,892 &  21,392 & 6,187 & 29,049 &  7,556 & 265,448 & 6,394 & 3,355 & 12,062 &  16,095 & 30,023 &  424,453  \\\hline 
    Preprocessed (Unique)& 25,498 &  20,240 & 5,764 & 26,514 &  7,263 & 248,580 & 6,027 & 3,226 & 11,658 &  15,346 & 27,223 &  397,423 \\\hline 
    \end{tabular}
    }
    \label{tab:tweet_statistics}
\end{table*}


The tweets are then pre-processed by expanding contractions, removing the user mentions, reserved words (i.e., RT), emojis, smileys, numeric tokens, URLs, HTML tags. 
The punctuation marks are also removed. Moreover, the tokens except hashtags are lemmatized. Eventually, the ``\#" in hashtags are removed, while the tokens in hashtags are reserved. Finally, the stop-words are removed and only the tweets of sentence length greater than 1 are retained. More specifically for topic modeling (cf. Section~\ref{sec:topic_modeling}), words with document frequency above 70\% are removed. Table~\ref{tab:tweet_statistics} shows the statistics of the extracted and preprocessed tweets.

\subsection{Topic Modeling}
\label{sec:topic_modeling}

In the previous step, the tweets are collected based on keywords, so it is inevitable that a lot of tweets are actually irrelevant to the topic of migration. For example,
many tweets including the keyword ``migrant” are about the topic ``migrant birds”, and the tweets containing the keyword ``exile” are about the ``Japanese Band Exile". Due to the large number of tweets collected (i.e., 397,423 preprocessed tweets), it is hard to manually filter out irrelevant tweets. In order to automate this process, topic modeling is performed.

Topic Modeling is used for extracting hidden semantic structures in the textual documents. One of the classical algorithms for topic modeling is Latent Dirichlet Allocation (LDA)~\citep{BleiNJ03} which represents each topic as the distribution over terms and each document as a mixture of topics. It is a very powerful algorithm, but it fails in case of huge vocabulary. Therefore, for the current study, the most recent topic modeling algorithm, Embedded Topic Model (ETM)~\citep{DiengRB20}, is chosen. Similar to LDA, ETM models each document as a mixture of topics and the words are generated such that they belong to the topics (ranked according to their probability). It also relies on topic proportion and topic assignment. Topic proportion is the proportion of words in a document that belong to a topic, which are the main topics in the document. Topic assignment refers to important words in a given topic. In addition to that, ETM makes use of the embedding of each term and represents each topic in that embedding space. In word embeddings, the context of the word is determined by its surrounding words in a vector space but in case of ETM the context is defined based on the topic. The topic’s distribution over terms is proportional to the inner product of the topic’s embedding and each term’s embedding.


In the setting of the current study, the word and the topic embeddings are trained on tweets. First, the word embeddings are generated by training a Word2Vec skip-gram model on all the preprocessed tweets for 20 epochs, with minimal word frequency 2, dimension 300, negative samples 10 and window size 4. For obtaining the optimal training parameters for ETM, its performance is computed on a document completion task~\citep{rosen-zvi2004, wallach2009}. The parameters for which the highest performance is achieved, are selected and consequently ETM model is utilized. In order to obtain optimal parameters, the dataset is split into 85\%, 10\%, 5\% for train, test, and validation set respectively. 
The size of the vocabulary of the dataset is 22850. To explore the optimal number of topics, the ETM is experimented with 25, 50, 75, and 100 topics. Initialized with the pretrained word embeddings, the ETM is trained on training data, with batch size 1000, Adam optimizer, and ReLU activation function. In order to select the best number of epochs for training ETM, the model is trained repeatedly by selecting 1-200 epochs and evaluated on the task of document completion (as described previously). The model performs the best on 172 number of epochs with 50 topics.

\begin{table}[htb]
    \centering
        \caption{The Results of ETM with Different Number of Topics. Bold values show the best results.}

    \begin{tabular}{|l|c|c|c|c|}
    \hline 
      \textbf{Nr. Topics}   & \textbf{25} & \textbf{50} & \textbf{75} & \textbf{100}   \\\hline
        Val PPL & 3329 & 3015 & 2920 & \textbf{2870} \\\hline 
        Best Epoch & 185 & 172 & 176 & 178 \\\hline 
        Topic Coherence & 0.0744 & \textbf{0.0777} & 0.0506 & 0.02 \\\hline
        Topic Diversity & \textbf{0.9696} & 0.9288 & 0.9056 & 0.7832 \\\hline 
        Topic Quality & \textbf{0.0721} & \textbf{0.0721} & 0.0460 & 0.0157 \\\hline 
        Classfied Nr. Of Topics & 25 & 50 & 75 & 87 \\\hline 
    \end{tabular}
    \label{tab:etm_results}
\end{table}

    
    
    


The metrics topic coherence and topic diversity are used for evaluation~\citep{DiengRB20}. Topic coherence provides a quantitative measure of the interpretability of a topic~\citep{mimno2011}, which is the average point-wise mutual information of two words drawn randomly from the same tweet. A coherent topic would display words that are more likely to occur in the same tweet. In turn, the most likely words in a coherent topic should have high mutual information. In contrary, the topic diversity is defined as the percentage of unique words in the top 25 words of all topics. If there are topics that contain high percentage of words that overlap with the words in another topic, i.e., the diversity would be low, then the topics are redundant. If the diversity is close to 1, the topics are diverse. The results for models with different number of topics are shown in Table~\ref{tab:etm_results}. The model with 100 topics has the lowest topic diversity and topic coherence, and only 87 topics are assigned to the tweets, which indicates redundancy of the topics. Comparing the models with 25, 50 and 75 topics, the model with 50 topics has a comparably the best topic quality and provides a wider range of topics. Therefore, the trained ETM model with 50 topics is used for classifying the topics for the tweets. 





\begin{table}[htb]
    \centering
    \caption{Example of topic, terms belonging to the topics, an example Tweet, and its maximal probability score regarding the migration-related topics. The topics with * are chosen as migration-related. } 
\resizebox{\textwidth}{!}{
    \begin{tabular}{|p{1cm}|p{5.5cm}|p{10cm}|p{2cm}|}
    \hline 
       \textbf{Topic}  &  \textbf{Top Words} & \textbf{Preprocessed Tweet} & \textbf{Max. Probability Score} \\\hline
       
       1* & refugee, seeker, kill, alien, enter & treatment refugee violate human rights dehumanize refugee endanger european value security argue group psychologist open letter &  0.7195 \\\hline 
       2* & great, call, immigration, question, town & peddle lie interwoven thread brexit regional leave voter low exposure immigration easy scare foreigner queue town come assimilate quickly & 1.1062 \\\hline 
       3* & work, refugee, covid, border, woman &  yeah let corrupt nhs education system fine cause deport load hard work immigrant & 0.8585 \\\hline 
       4 & people, take, uk, health, hope &  illegal immigrant get day uk free home cash health education maternity british national take fool katiehopkins & 0.9598 \\\hline 
       
       5 & stop, find, austria, future, country & proven liar self promote cheat allow uncounted unchecked immigration country cause current crisis & 0.4782 \\\hline 
        
    \end{tabular}
    }
    \label{tab:top_topic}
\end{table}

\begin{figure}[htb]
    \centering
        \caption{The distribution of maximal probability scores of tweets regarding migration-related topics before filtering. }

    \includegraphics[width=\textwidth]{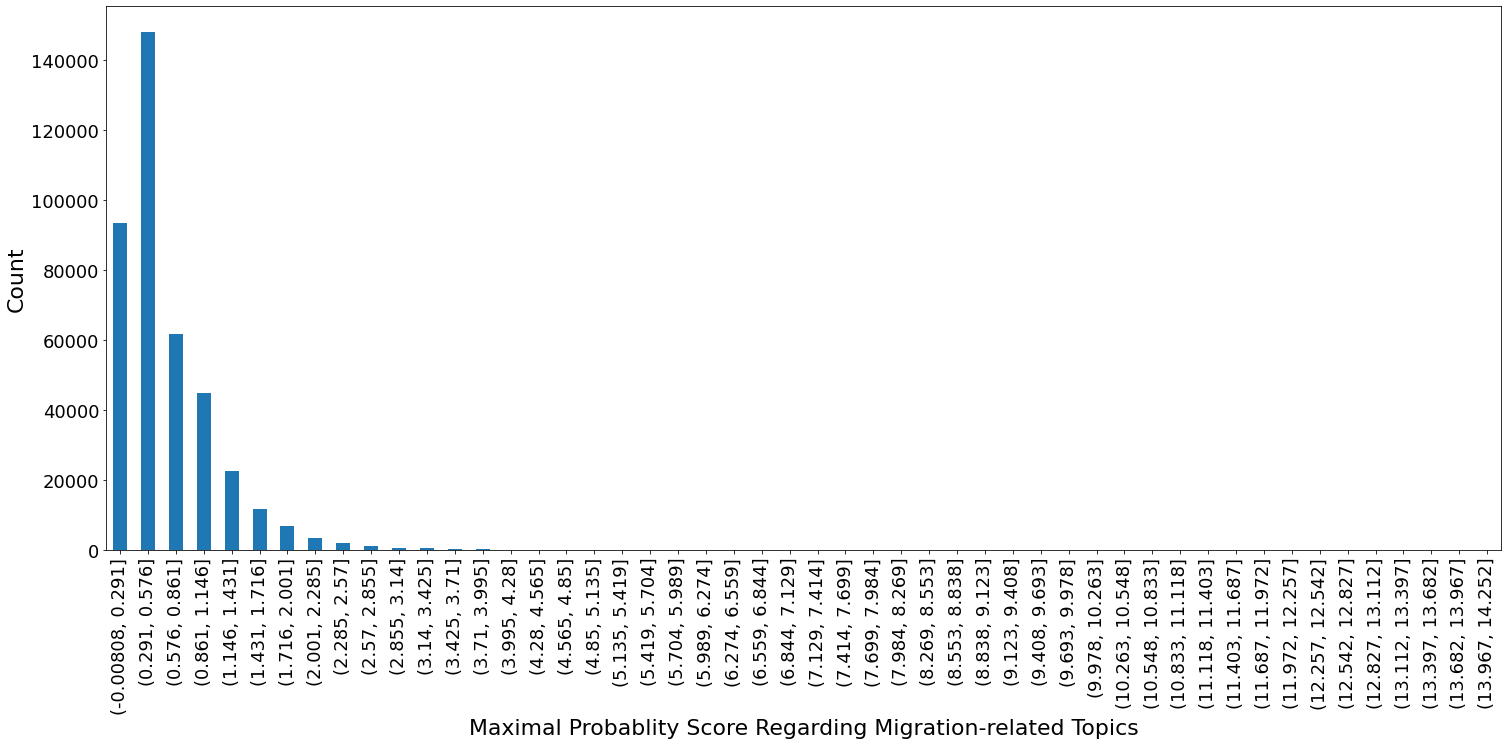}
    \label{fig:dist_max_score}
\end{figure}

\begin{figure}[htb]
    \centering
        \caption{The distribution of maximal probability scores of tweets regarding migration-related topics after filtering.}

    \includegraphics[width=\textwidth]{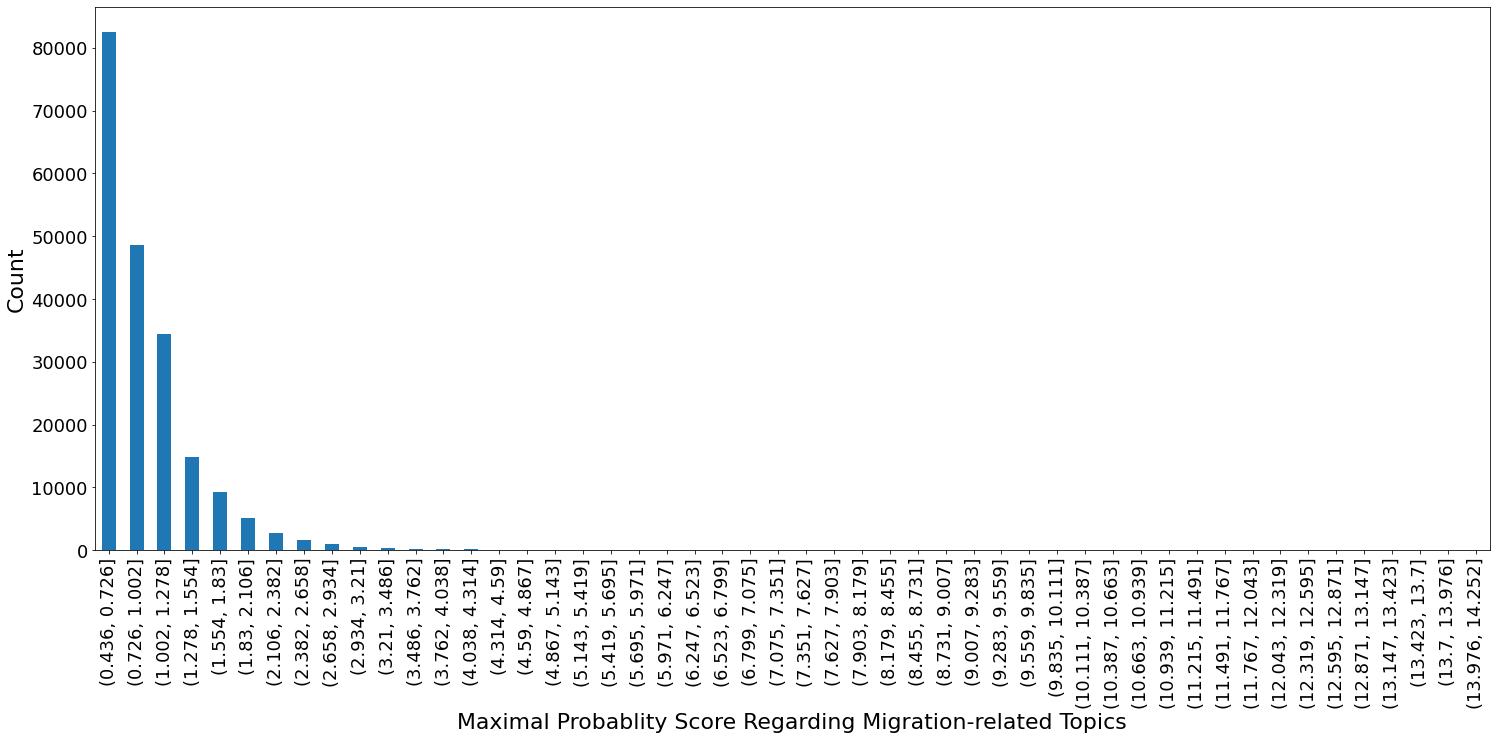}
    \label{fig:dist_max_score_t45}
\end{figure}

\begin{figure}[htb]
    \centering
    \caption{The distribution of topics before filtering. }
    \includegraphics[width=\textwidth]{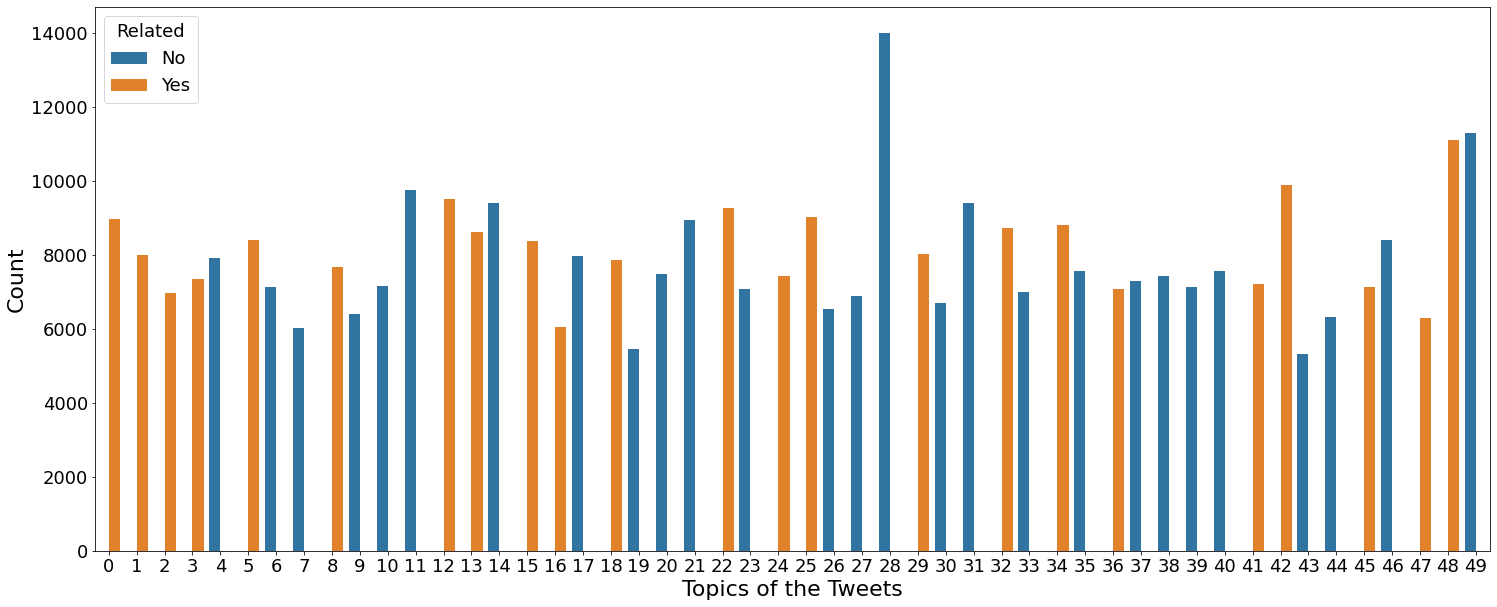}
    
    \label{fig:dist_topic_before_filtering}
\end{figure}

\begin{figure}[htb]
    \centering
    \caption{The distribution of topics after filtering. }
    \includegraphics[width=\textwidth]{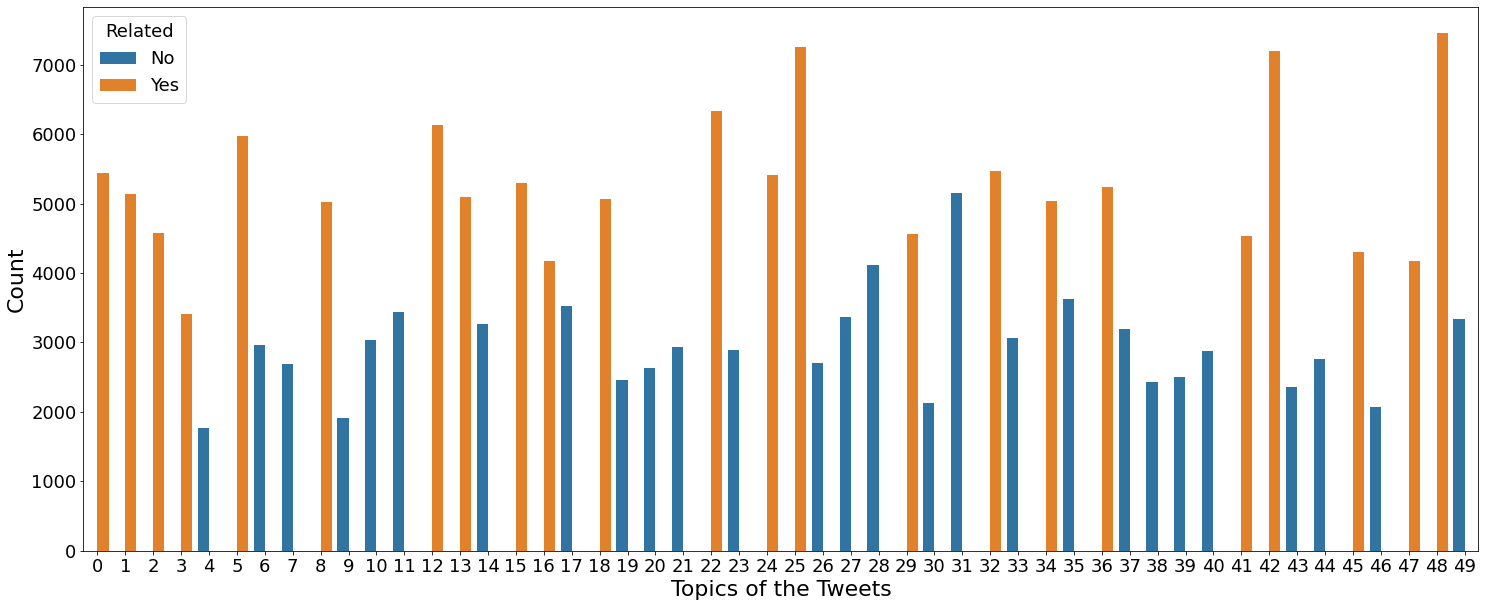}
    
    \label{fig:dist_topic_after_filtering}
\end{figure}

The tweets are then refined based on topic embeddings.
For each topic, the top 20 words (ranked by their probability) are selected as a representation of the topic\footnote{\url{https://bit.ly/2SZgpKb}}. These words are then manually verified based on their relevance to the topic of migration. The migration-related topics are chosen with the help of the probabilities associated to all the topics. Regarding the chosen topics, the maximal probability score for each tweet is extracted. Figure~\ref{fig:dist_max_score} shows the distribution of the maximal probability scores of the tweets regarding the chosen topics. 
By manual evaluation, the threshold for reserving the tweets by the probability score is set to 0.45. Figure~\ref{fig:dist_max_score_t45} shows the distribution of the probability scores of tweets after filtering over the threshold. Eventually, out of the original 397423 tweets, 201555 are reserved for further analysis. Moreover, the topic of each tweet is defined by the maximal probability from all the topics. For example, in Table~\ref{tab:top_topic}, the topics 1,2, and 3 belong to the chosen migration-related topics, while 4 and 5 are not. More specifically, for the tweet ``illegal immigrant get day uk free home cash health education maternity british national take fool kaitehopkins", it is classified as topic 4, and has the maximal probability score 0.9598, which is over the threshold 0.45 and is reserved for the {\tt MigrationsKB}. As shown in Figure~\ref{fig:dist_topic_before_filtering} and Figure~\ref{fig:dist_topic_after_filtering}, the tweets are more evenly distributed before filtering, while there are more tweets with migration-related topics proportionally after filtering.

\begin{figure*}[htb]
    \centering
        \caption{Distribution of all the Crawled Tweets based on geographic location.}

    \includegraphics[width=0.49\textwidth]{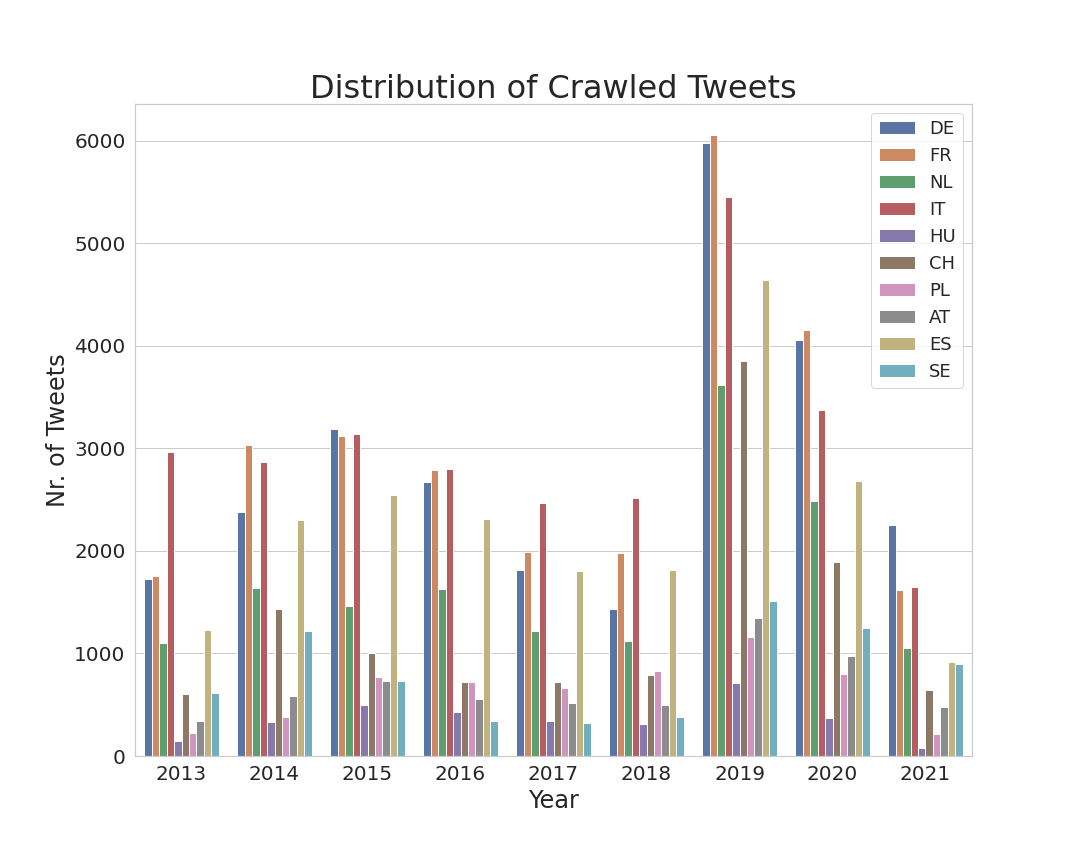}
    \includegraphics[width=0.49\textwidth]{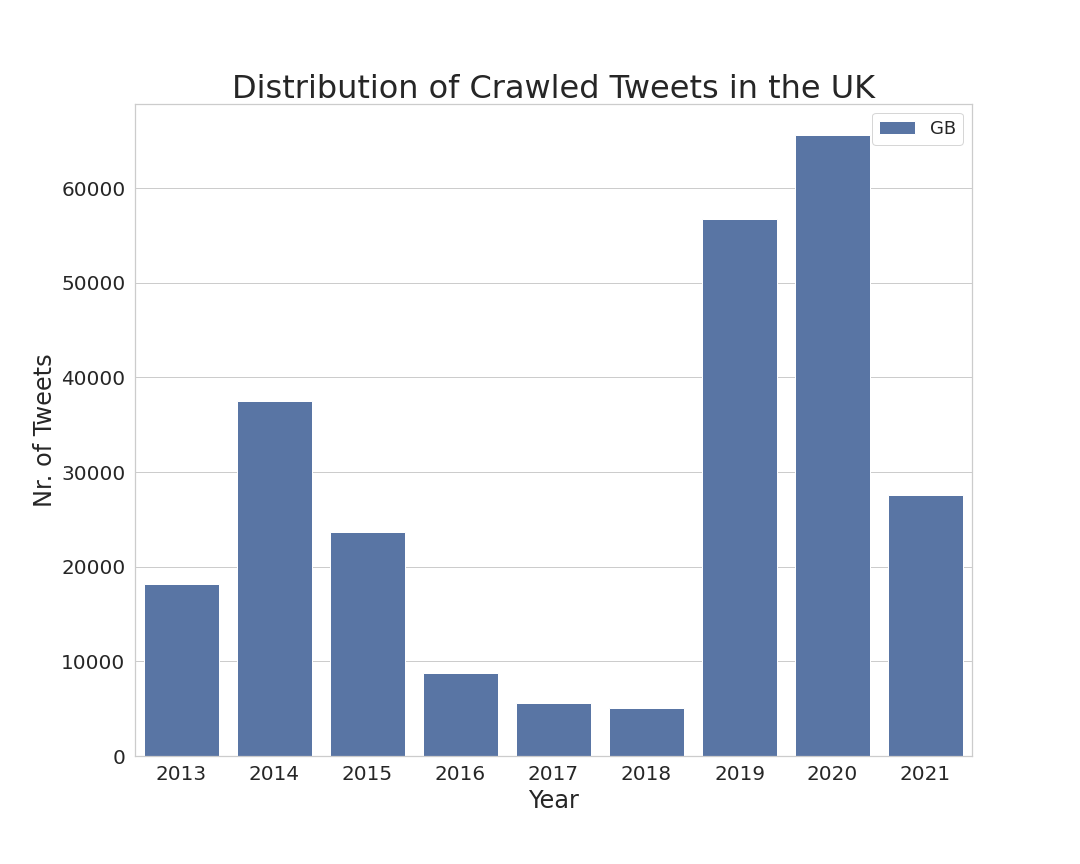}
    
    \label{fig:dist_before_etm}
\end{figure*}

\begin{figure*}[htb]
    \centering
        \caption{Distribution of Tweets based on geographic location after filtering using ETM.}

    \includegraphics[width=0.49\textwidth]{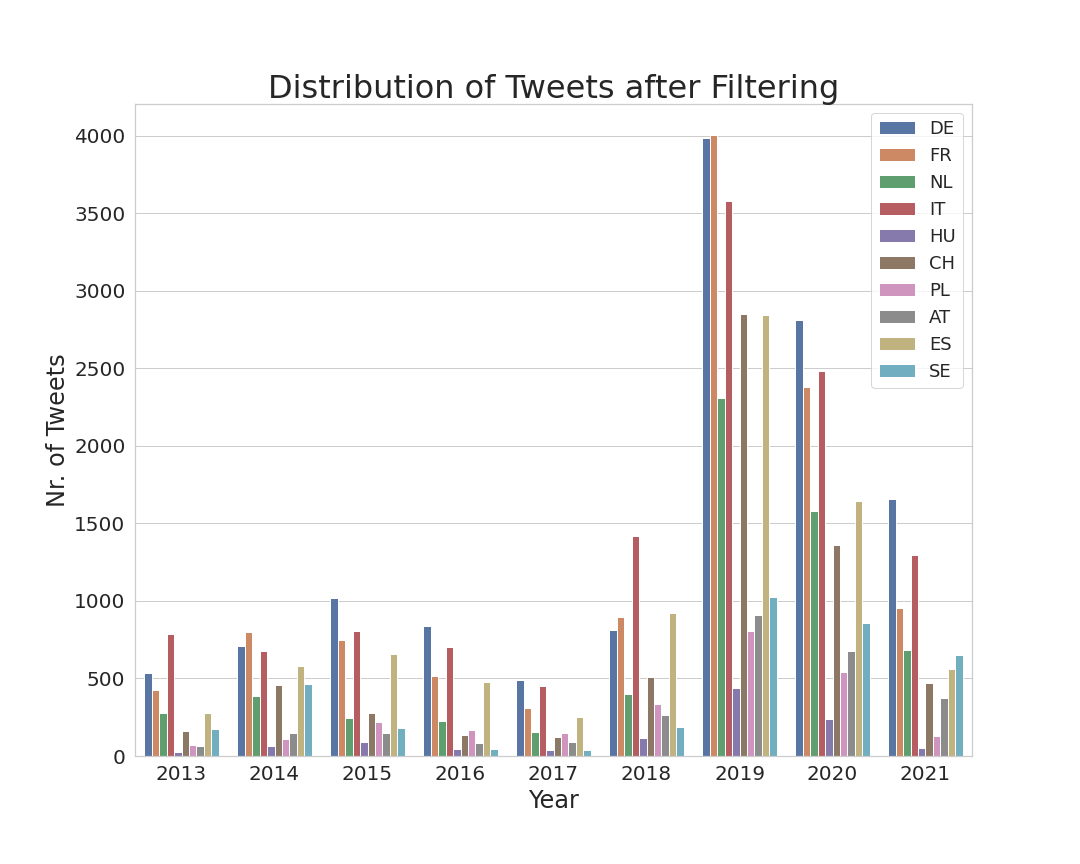}
    \includegraphics[width=0.49\textwidth]{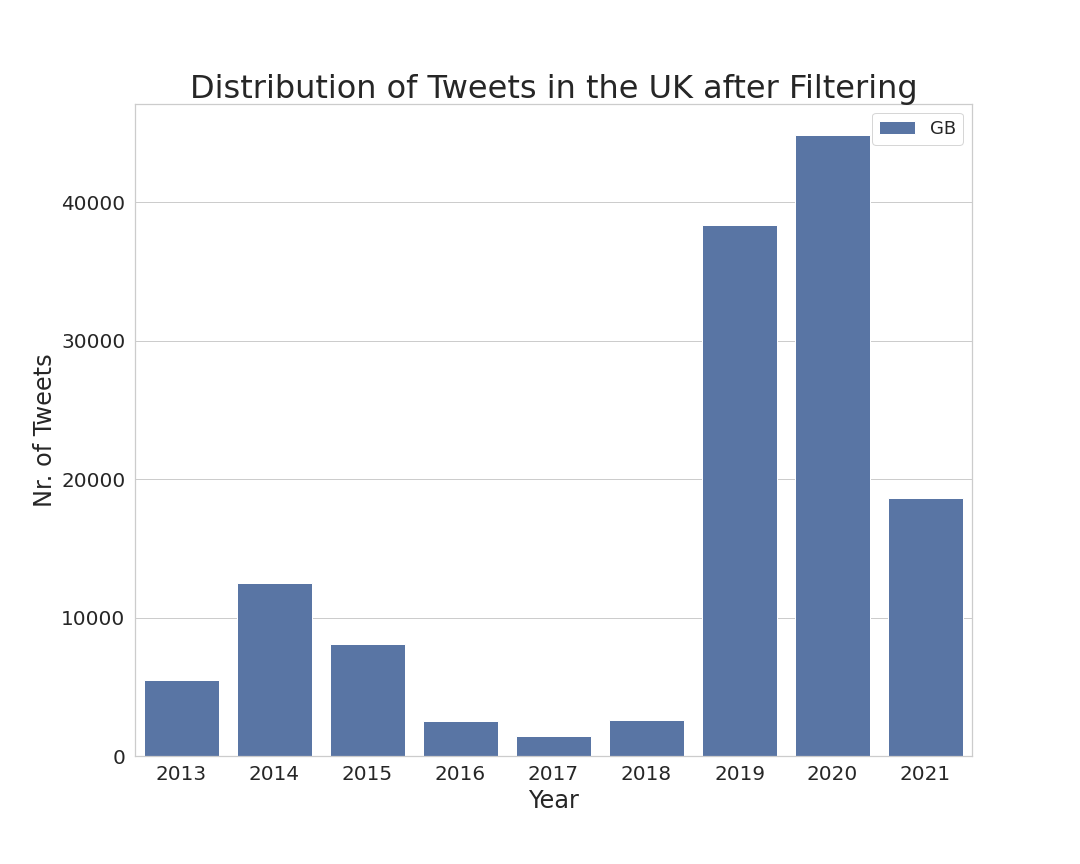}
    \label{fig:dist_after_etm}
\end{figure*}

Figure~\ref{fig:dist_before_etm} (left) shows the distribution of all the crawled tweets from destination countries from 2013 to Jul-2021. Most of the tweets are from the year 2019 geotagged with Germany, France, Netherlands, Italy, and Spain. Figure~\ref{fig:dist_before_etm} (right) shows the distribution of all the crawled tweets with the geotag UK within the time frame from 2013 to Jul-2021. Most of the tweets are from the year 2019 and 2020. UK is chosen because currently the focus of this study is English language and the majority of tweets are from there. Figure~\ref{fig:dist_after_etm} shows the distribution of the filtered tweets after using ETM. For all the countries, the graph shows similar proportions/trends as Figure~\ref{fig:dist_before_etm}, but the number of tweets is obviously lower.


\subsection{Sentiment Analysis}

In order to measure the public attitudes towards migrations, sentiment analysis is performed by classifying the tweets into positive, negative, and neutral sentiments. These public sentiments in the destination countries are then analyzed based on their geographic location and temporal information.

\paragraph{Training Dataset Selection.} 
Since, there is a lack of datasets available for sentiment analysis particular to the domain of migration, the existing twitter datasets for sentiment analysis are used for fine-tuning language models for transfer learning on the collected data. Two twitter datasets for sentiment analysis are used most frequently, i.e., the Airline dataset \footnote{\url{https://bit.ly/3u49hZT}} and the SemEval2017 dataset \footnote{\url{https://bit.ly/3v2jUOd}}\citep{rosenthal-etal-2017-semeval}. The Airline dataset focuses on travelers' opinions on Twitter, which is domain specific. In comparison, the SemEval2017 dataset consists of broader topics of tweets including a range of named entities (e.g., Donald Trump, iPhone), geopolitical entities (e.g., Aleppo, Palestine), and other entities (e.g., Syrian refugees, gun control, and vegetarianism). The dataset is manually annotated using CrowdFlower. 
The language models are fine-tuned on both the datasets separately and on the combination. Table~\ref{tab:senti_datasets} shows the statistics of the datasets.



\begin{table}[htb]
    \centering
    \caption{Statistics of the Existing Datasets for Sentiment Analysis.}
    \begin{tabular}{|p{2cm}|p{1.5cm}|p{1.5cm}|p{1.5cm}|p{1cm}|}
    \hline 
      &  \textbf{Train} & \textbf{Validation} & \textbf{Test } & \textbf{All} \\\hline 
            \multicolumn{5}{|l|}{\bf SemEval2017} \\\hline 

      negative  &   6291 &	752	 & 766 &	7809  \\\hline 
      neutral & 17981 &	2256 & 2287 & 	22524 \\\hline 
      positive & 15833 & 2006 & 	1960 & 	19799 \\\hline 
      \multicolumn{5}{|l|}{\bf Airline} \\\hline 
     
      negative  & 7316	& 923 & 939 & 9178 \\\hline 
      neutral & 2475 &	316 & 308 &	3099 \\\hline 
      positive & 1921 &	225 &	217 & 	2363 \\\hline 
        \multicolumn{5}{|l|}{\bf Combined} \\\hline 
      negative  & 13608	& 1736 & 1643 &	16987     \\\hline 
      neutral & 20516 &	2535 &	2572 &	25623    \\\hline 
      positive &   17693 &	2207 &	2262 &	22162     \\\hline
    \end{tabular}
    \label{tab:senti_datasets}
\end{table}

\paragraph{Model Selection for Sentiment Analysis.} 

For transfer learning, three contextual embedding models are chosen, i.e., BERT~\citep{DevlinCLT19}, XLNet~\citep{YangDYCSL19}, and ULMFit~\citep{RuderH18}. These three models are fine-tuned as mentioned earlier. The fine-tuned models are then tested on the test set of their corresponding datasets, as well as on the test set of the other datasets. The performance of each of these models is measured. Because the curated tweets obtained from previous steps are not domain specific, the fine-tuned language model is required to perform well on a non-domain specific dataset. Therefore, all the models are evaluated on the test set of SemEval2017 dataset. The results of all the models are shown in Table \ref{tab:sent_results}. As shown in the Table \ref{tab:senti_datasets}, there are more neutral and positive tweets than negative ones in SemEval2017 which leads to class imbalance. The macro metrics are more robust to class imbalance and reflect the real performance classifying the minority classes compared to micro metrics, hence the macro F1 score, macro precision, macro recall and also standard accuracy are reported. Since, BERT fine-tuned on SemEval2017 training dataset renders the best results, which is also the state-of-the-art on this dataset \citep{rosenthal-etal-2017-semeval}, it is chosen for transfer learning on the collected data for sentiment analysis.


\begin{table}[htb]
    \centering
    \caption{The Results of Contextual Embedding Models on SemEval2017 test dataset. Bold values show the best results.}
    \begin{tabular}{|c|c|c|c|c|c|}
    \hline 
      \textbf{Model}  &{\bf Fine-tuned}  & {\bf Acc} &  {\bf F1}  & {\bf Prec} & {\bf Rec} \\\hline 
      
      XLNet   & SemEval2017 & 0.7066 & 0.6851 & 0.6988	& 0.6719 \\\hline 
      XLNet & Airline & 0.5565	 & 0.5987 & 	0.5965	& 0.601 \\\hline 
      XLNet   & -  & 0.3718 & 	0.3482	& 0.3348	& 0.3627 \\\hline 
      BERT & SemEval2017 & \textbf{0.7068} &	\textbf{0.6949} & \textbf{0.7007} &	\textbf{0.6892}  \\\hline
       ULMFiT & SemEval2017 & 0.6624 &	0.6365 & 0.6342 & 0.6388 \\\hline 
      ULMFiT & Airline & 0.4709 & 	0.5215	& 0.52 & 	0.5231 \\\hline
     
      BERT & Airline & 0.5117 &	0.5831 & 	0.5736 & 	0.5929 \\\hline
      BERT & - & 0.5417 &	0.5722 & 	0.5753 & 	0.5692 \\\hline
      BERT & Combined & 0.6691 & 	0.6627 & 	0.6484 &	0.6776 \\\hline 
      
    \end{tabular}
    
    \label{tab:sent_results}
\end{table}

\paragraph{Analyzing Public Sentiments Towards Migrations.}

\begin{figure*}[htb]
    \centering
     \caption{Temporal Distribution of the Sentiments of the Public towards Migrations. The left image shows the sentiments of the people towards migrations in the United Kingdom, and the right image shows the sentiments for all 11 destination countries in Europe.}
    \includegraphics[width=0.49\textwidth]{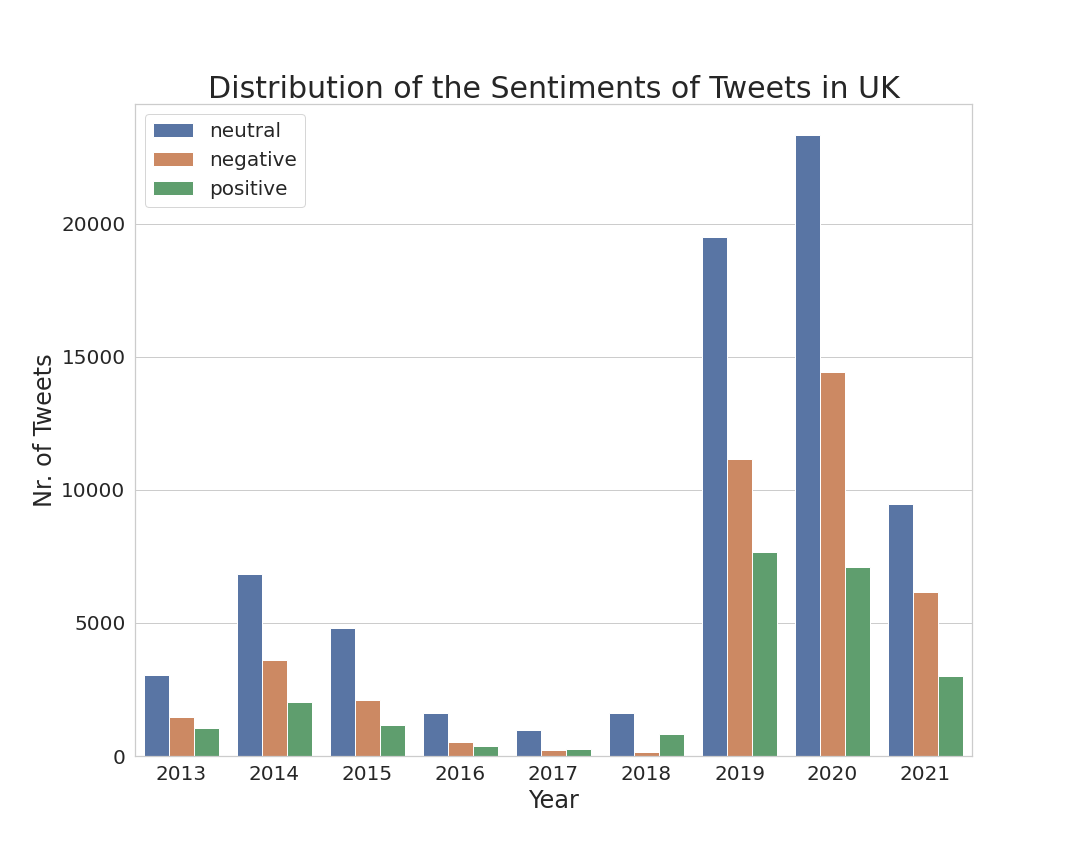}
    \includegraphics[width=0.49\textwidth]{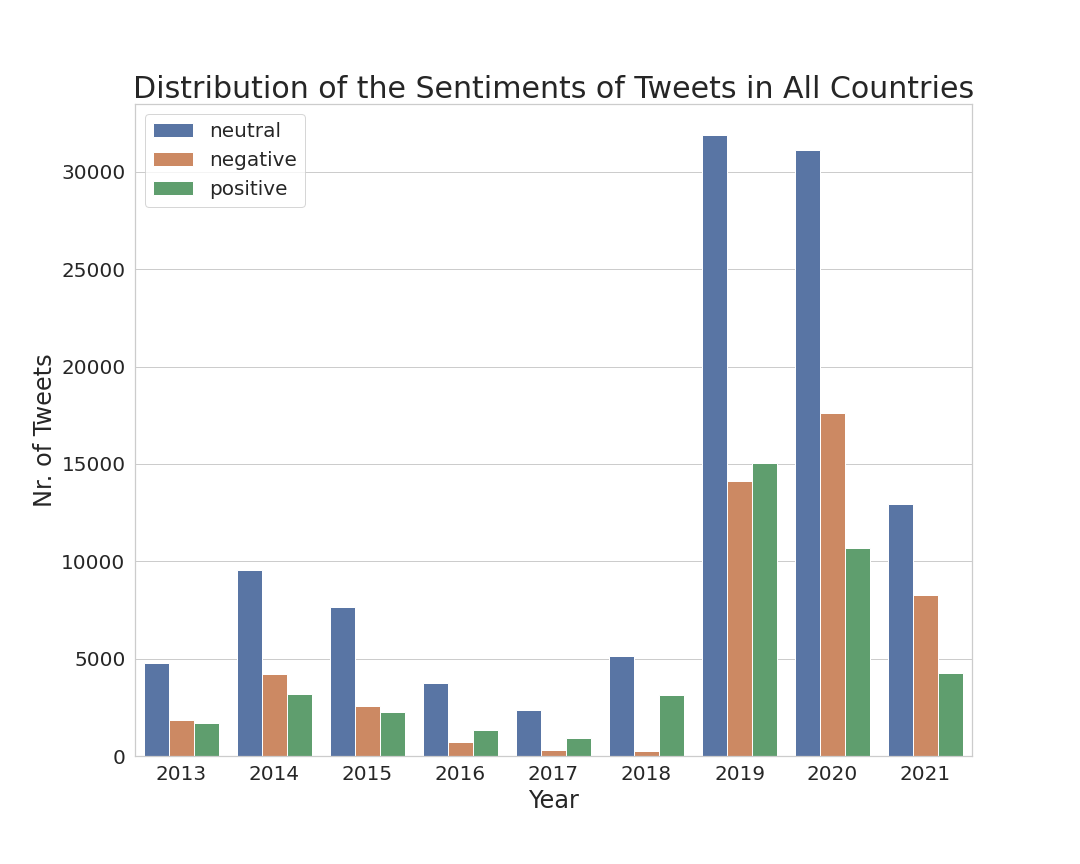}
    \label{fig:senti}
\end{figure*}

In order to identify the public attitudes towards migrations, the sentiments of the tweets for each country in each year are aggregated. In Figure~\ref{fig:senti} (left), the public sentiments in the UK from 2013 to Jul-2021 are shown. It can be observed that,  the total number of tweets about migration from 2013 to 2014, and the tweets with both positive and negative sentiments are increasing in the similar proportion. From 2016 to 2018, the topic of migration is less popular, while there is again a sharp increase in the tweets about migration from 2018 to 2020. Overall, the negative sentiment is much more significant towards migrations compared to the positive sentiment. As shown in Figure~\ref{fig:senti} (right), the public sentiment towards migrations in all 11 destination European countries follow similar trends as in the UK from 2013 to Jul-2021.

\subsection{Hate Speech Detection}

To measure the negative attitude of the public towards migrations, hate speech detection is performed. The tweets are classified into one of the three classes hate, offensive, and normal. 
In order to perform transfer learning in this scenario, all the hate speech detection models are trained on recently published manually annotated data about Hate Speech Detection, called as HateXplain~\citep{Mathew21}. Similar to previous studies on Hate Speech Detection, the sources of the dataset are Twitter \citep{waseem-hovy2016, davidson2017, founta2018} and Gab~\citep{mathew2019}. For HateXplain Twitter dataset, the tweets are filtered from the 1\% randomly collected tweets from Jan-2019 to Jun-2020 using the lexicons combined from \citep{davidson2017}, \citep{ousidhoum}, and \citep{mathew2019a}. The Gab dataset is originally introduced in \citep{mathew2019a}. All the data is annotated using Amazon Mechanical Turk (MTurk) where each text is annotated based on: (1) whether it is hate speech, offensive speech or normal; (2) the target communities in the text, including target groups such as Race, Religion, Gender, Sexual Orientation, and Miscellaneous; (3) if the text is considered as hate speech or offensive speech by majority, the annotators further annotate which parts of the text provide rationale for the given annotation (this ensures the explainability of manual annotation by the annotators).

HateXplain is split into train, validation, and test dataset by 80\%, 10\%, and 10\%, for which the stratified split is performed to maintain class balance. BiRNN \citep{birnn} and BiRNN-Attention \citep{birnn-attn} are widely used for text classification task, and CNN-GRU\citep{zhang2018} is used for hate speech detection. In the current study, the experimentation is conducted using a combination of various models from CNN, BiLSTM, GRU, BiGRU, and an attention layer for selecting the best model for hate speech detection. For all the models, pre-trained on GloVe embeddings~\citep{glove2014} are used as reported in \citep{Mathew21}. A dropout layer with dropout rate 0.3 is applied after the word embedding layer. For CNN models the convolutional layer has a filter size 100, and window sizes 2, 3, 4. The RNN models use hidden size 100. Finally, the softmax function is used for classifying the texts. The models with the highest accuracy on validation dataset (after training on the training dataset) are chosen for test on the test dataset, whose results are reported in Table~\ref{tab:hate_speech}. Eventually, the best performing pretrained model CNN+BiLSTM+Attention, which is comparable to the results of the best performing model from \citep{Mathew21}, is used for transfer learning on the collected tweets.


\begin{table}[!htb]
    \centering
    \caption{The statistics of the HateXplain Dataset and the results of different Hate Speech Detection models.}
    \begin{subtable}{.3\linewidth}
    \caption{The Statistics of HateXplain Dataset.}
    \begin{tabular}{|p{1.2cm}|p{1cm}|p{1.2cm}|p{1cm}|}
    \hline
     
     {\bf Dataset} & {\bf Normal}   & {\bf Offensive}  & {\bf Hateful} \\\hline
      Train   &  6251 & 4384  & 4748   \\\hline
      Validation & 781 & 548 & 593  \\\hline
      Test  & 782  & 548 &  594  \\\hline 
    \end{tabular}
    \end{subtable} \hfill
    \begin{subtable}{.6\linewidth}
       \caption{The Results of Hate Speech Models on HateXplain. Bold values show the best results.}
        \begin{tabular}{|l|c|c|c|c|}
        \hline 
       \textbf{ Model}           & \textbf{Acc}   & \textbf{F1}    & \textbf{Prec}   & \textbf{Rec}  \\\hline 
        BiGRU	 & 0.6533 & 	0.6353	& 0.6343 & 	0.6364 \\\hline 
        BiGRU+Attn	 & 	0.6445 & 	0.6344 & 	0.6297 & 	0.6392 \\\hline 
         BiLSTM		& 0.6284 & 	0.6211 & 	0.6169 & 	0.6253 \\\hline 
        BiLSTM+Attn	 &  0.6512 & 	0.6421 & 	0.6386 & 	0.6457 \\\hline 
        
        CNN+GRU		 & 	0.6544	&  0.6545 &  0.6541  & 	0.6549 \\\hline 
        CNN+GRU+Attn & 	0.6450 & 	0.6330 & 	0.6372 & 	0.6436 \\\hline 
    
        CNN+BiGRU	 & 0.6575 & 	0.6489 & 	0.6461 & 	0.6517 \\\hline 
        CNN+BiGRU+Attn	 & 	0.6606 & 	0.6472 & 	0.6444 & 	0.6501 \\\hline 
         CNN+BiLSTM	 &	0.6372 & 0.6496	& 0.6523 &	0.647 \\\hline 
       \textbf{ CNN+BiLSTM+Attn}	&   \textbf{0.6863}	 &  \textbf{0.6751} & 	\textbf{0.6782} & 	\textbf{0.672} \\\hline 
    \end{tabular}
    \label{tab:hatexplain_dataset}
    \end{subtable}
    \label{tab:hate_speech}
    
\end{table}

    



\begin{figure}
    \centering
        \caption{Temporal distribution of tweets after hate speech detection. The left image shows the distribution of tweets from UK while the right image is for all the 11 EU countries.}

\includegraphics[width=0.49\textwidth]{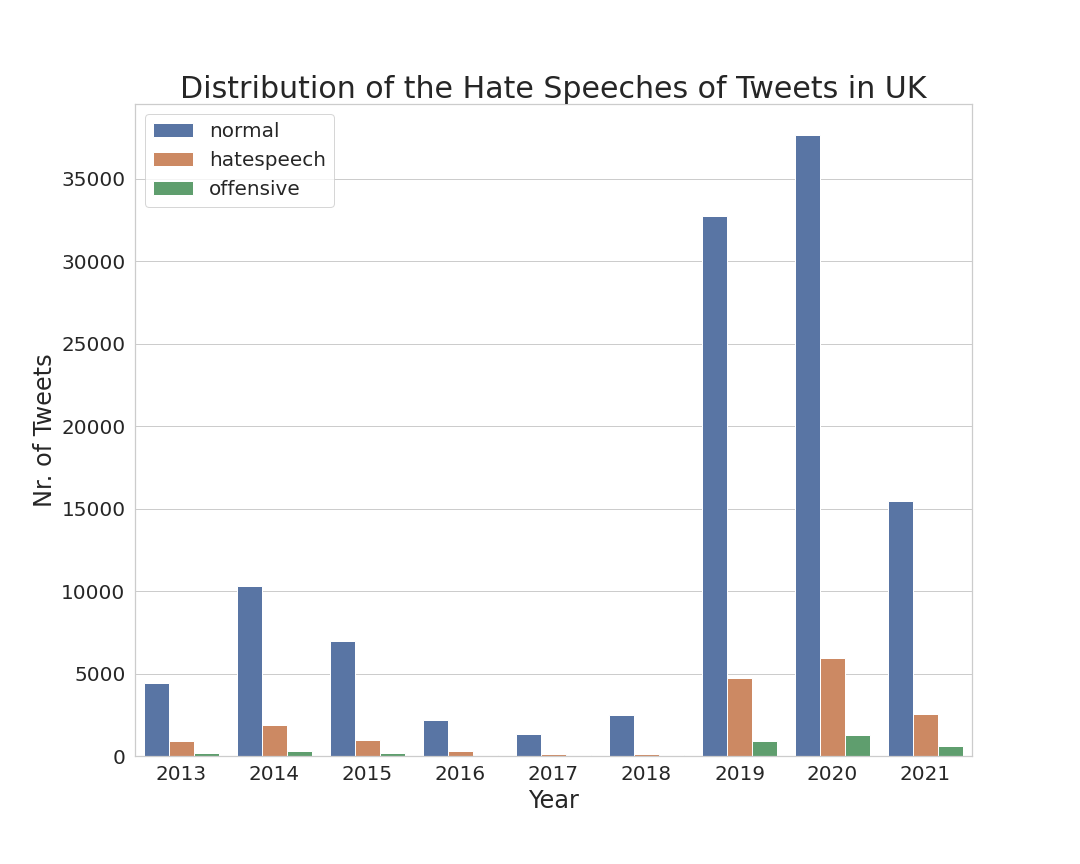}
    \includegraphics[width=0.49\textwidth]{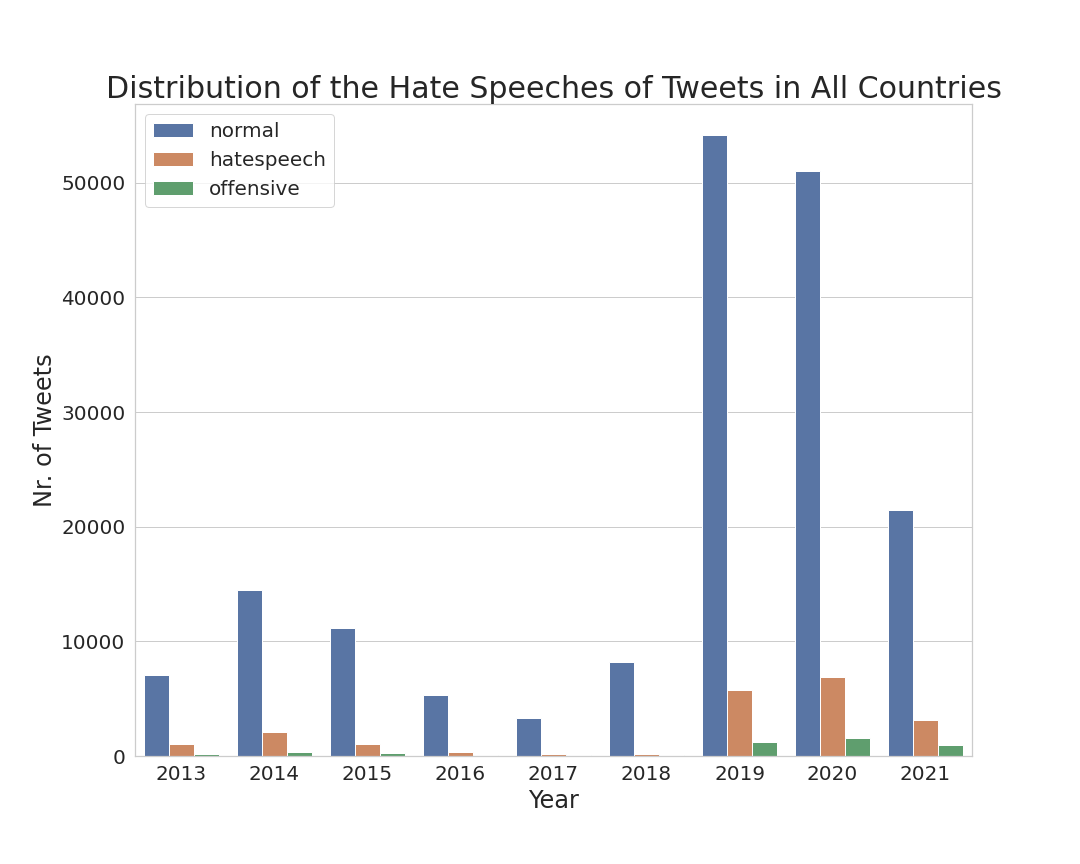}
    \label{fig:hate_speech_by_year}
\end{figure}

\paragraph{Analyzing Hate Speech in Public Opinions about Migration.}
The results for hate speech detection are aggregated temporally and geographically to identify the negative attitude of the public towards migrations. As shown in Figure~\ref{fig:hate_speech_by_year}, the number of hateful tweets is increasing from 2013 to 2014, decreasing from 2016 to 2018. It is then increasing again from 2019 to 2020 both in the UK and overall in 11 destination countries. In general, the proportion of offensive tweets and hateful tweets are always less than the tweets belonging to the normal class. In summary, the percentage of tweets classified as hate speech in the UK over 2013 to 2020 amounts to 12.98\%, while in 11 destination countries it is about 9.36\%.

\subsection{Entity Linking}

For entity linking BLINK~\citep{WuPJRZ20} is used, which utilizes Wikipedia\footnote{The 2019/08/01 Wikipedia dump, which is downloadable in its raw format from \url{http://dl.fbaipublicfiles.com/BLINK/enwiki-pages-articles.xml.bz2}}as the target KB.
Based on fine-tuned BERT, BLINK uses a two-stage approach. In the first stage, BLINK retrieves the candidates in a dense space defined by a bi-encoder that independently embeds the context of entity mention and the entity descriptions. Then in the second stage, each candidate is examined with a cross-encoder, that concatenates the entity mention and entity text. BLINK outperforms state-of-the-art methods on several zero-shot benchmarks and also on established non-zero-shot evaluations such as TACKBP-2010\footnote{\url{https://catalog.ldc.upenn.edu/LDC2018T16}}. Out of 201555 tweets in {\tt MigrationsKB}, for 145747 tweets there is at least one entity mention detected using BLINK. For one tweet, the maximum number of detected entity mentions is 30. In total 89076 unique entities are detected. The detected entities are available online\footnote{\url{https://bit.ly/3yPpa9D}}. The 10 most frequently detected entities containing ``refugee" is shown in Table~\ref{tab:entity_refugee}.

\subsection{Factors Effecting the Public Attitudes Towards Migrations}

\begin{figure*}[htb]
    \centering
    \caption{The trend of Hate Speech against immigrants/refugees in all the identified destination countries from 2013 to Jul-2021.}
    \includegraphics[width=0.8\textwidth]{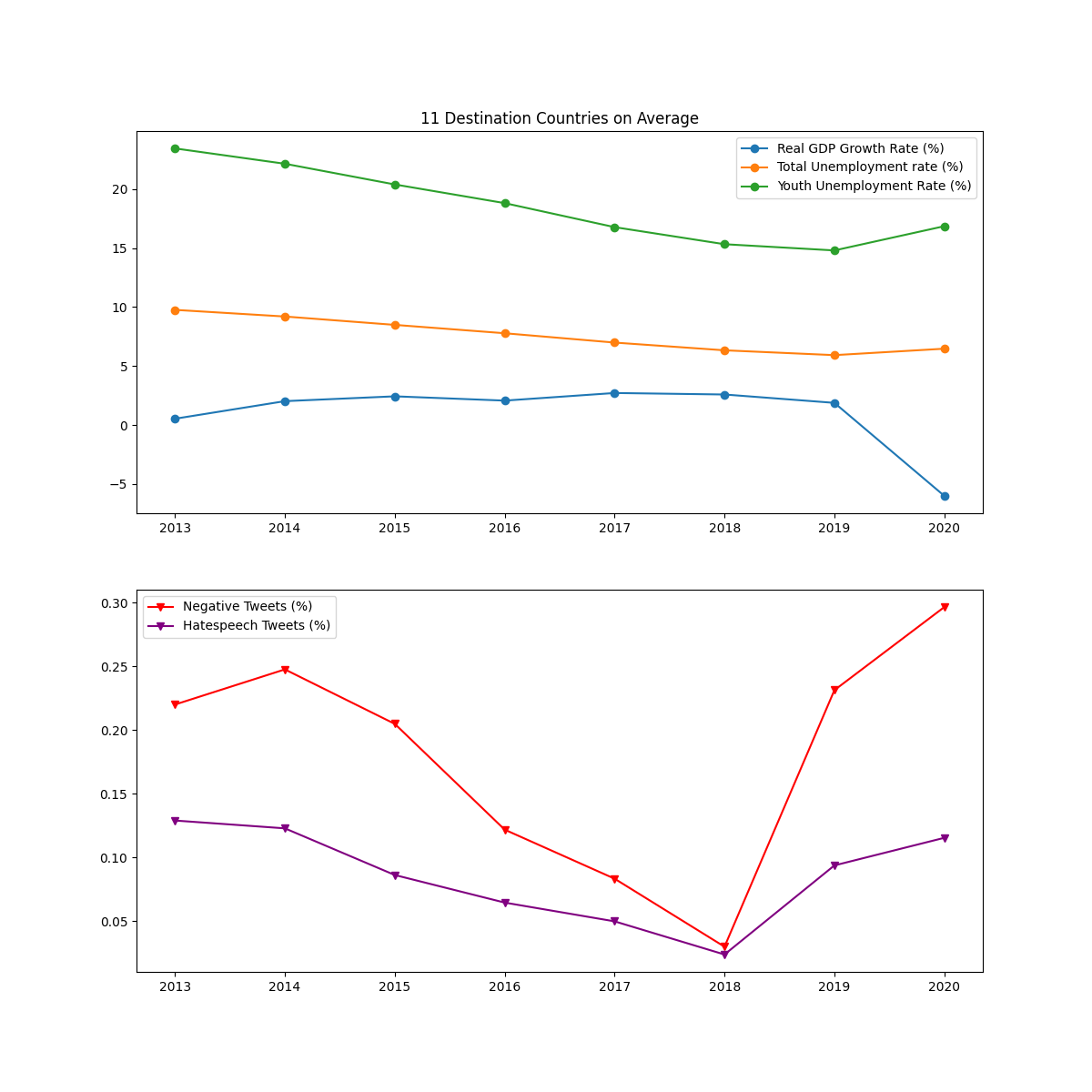}
    \label{fig:factors_all}
\end{figure*}

\begin{figure*}[htb]
    \centering
    \caption{The trend of hate speech against immigrants/refugees in the United Kingdom from 2013 to Jul-2021.}
    \includegraphics[width=0.8\textwidth]{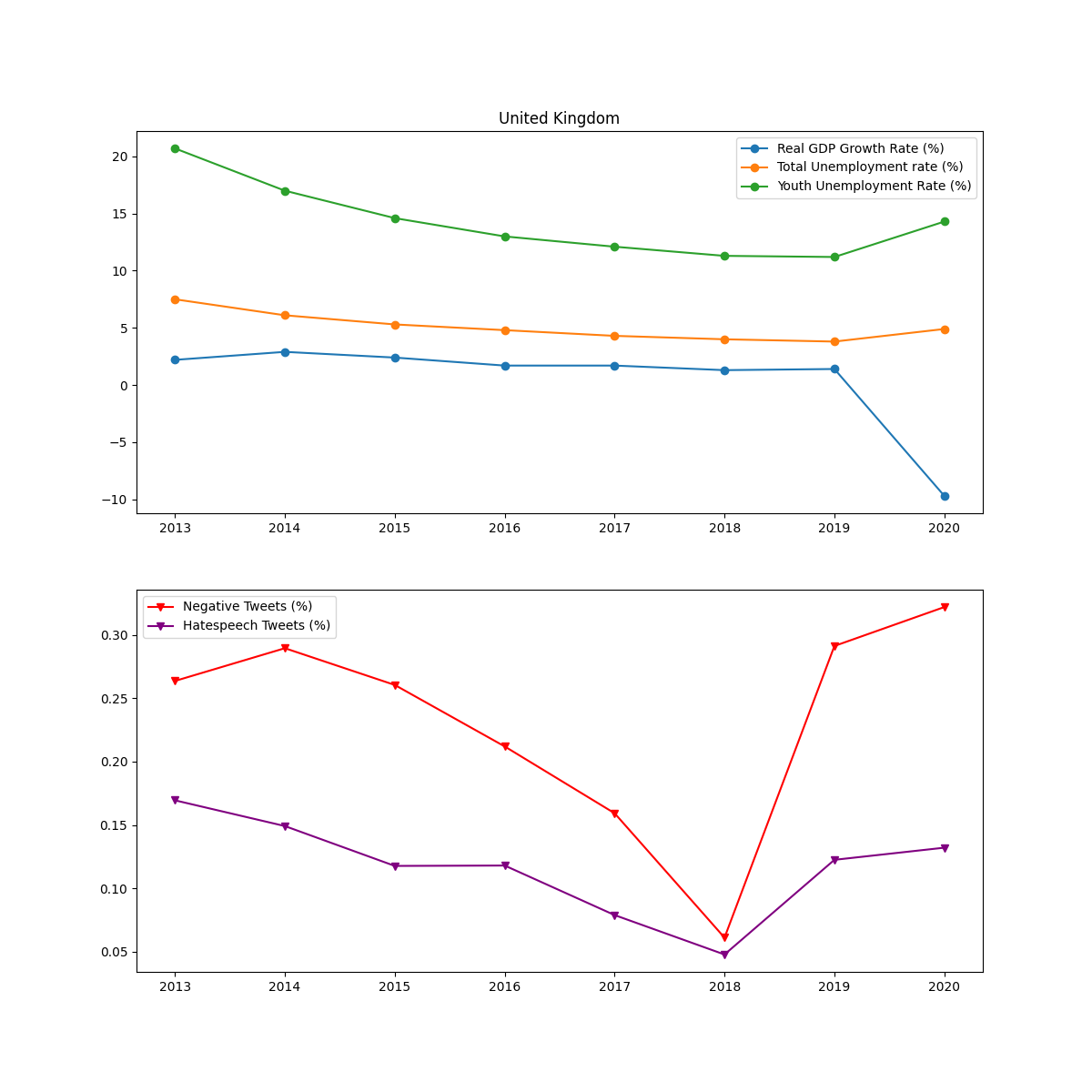}
    
    \label{fig:factors_de}
\end{figure*}

In order to learn the potential cause of the negative public attitudes towards migrations, the factors such as unemployment rate including youth unemployment rate and total unemployment rate, and GDPR are studied. These factors are identified by the experts~\citep{lenka2018} as the potential cause of negative attitude towards migrations. This data is collected from Eurostat, Statista, UK Parliament, and Office for National Statistics. 
Figure~\ref{fig:factors_all} shows the comparison between the factors (such as youth employment rate, total employment rate, and real GDPR) and the negative attitudes (i.e., negative sentiment and hate speeches) in all the extracted tweets.
On average in all 11 destination countries (see Figure \ref{fig:factors_all}) and individually in the UK (see Figure \ref{fig:factors_de}), the percentages of hate speech and negative sentiment of the public attitudes towards immigration are negatively correlated with the real GDPR and positively correlated with total/youth unemployment rate, from 2013 to 2018 and from 2019 to 2020. In 2019, the percentages of hateful tweets and negative tweets are rapidly increased by more than 2\% and 1\% respectively as compared to 2018. The analysis for each of 11 countries are posted online\footnote{\url{https://migrationskb.github.io/MGKB/stats}}.



 \section{{\tt MigrationsKB}}
\label{sec:MigrationsKB}

This section discusses the extensions in the RDF/S model of TweetsKB for incorporating public attitudes towards migrations, as well as the economic indicators which drive these attitudes. Moreover, it also discusses some scenarios and competency questions (also translated to SPARQL) that could be asked by the experts to the {\tt MigrationsKB}.

\subsection{RDF/S Model of {\tt MigrationsKB}}

\begin{figure*}[htb]
    \centering
    \includegraphics[width=0.7\textwidth]{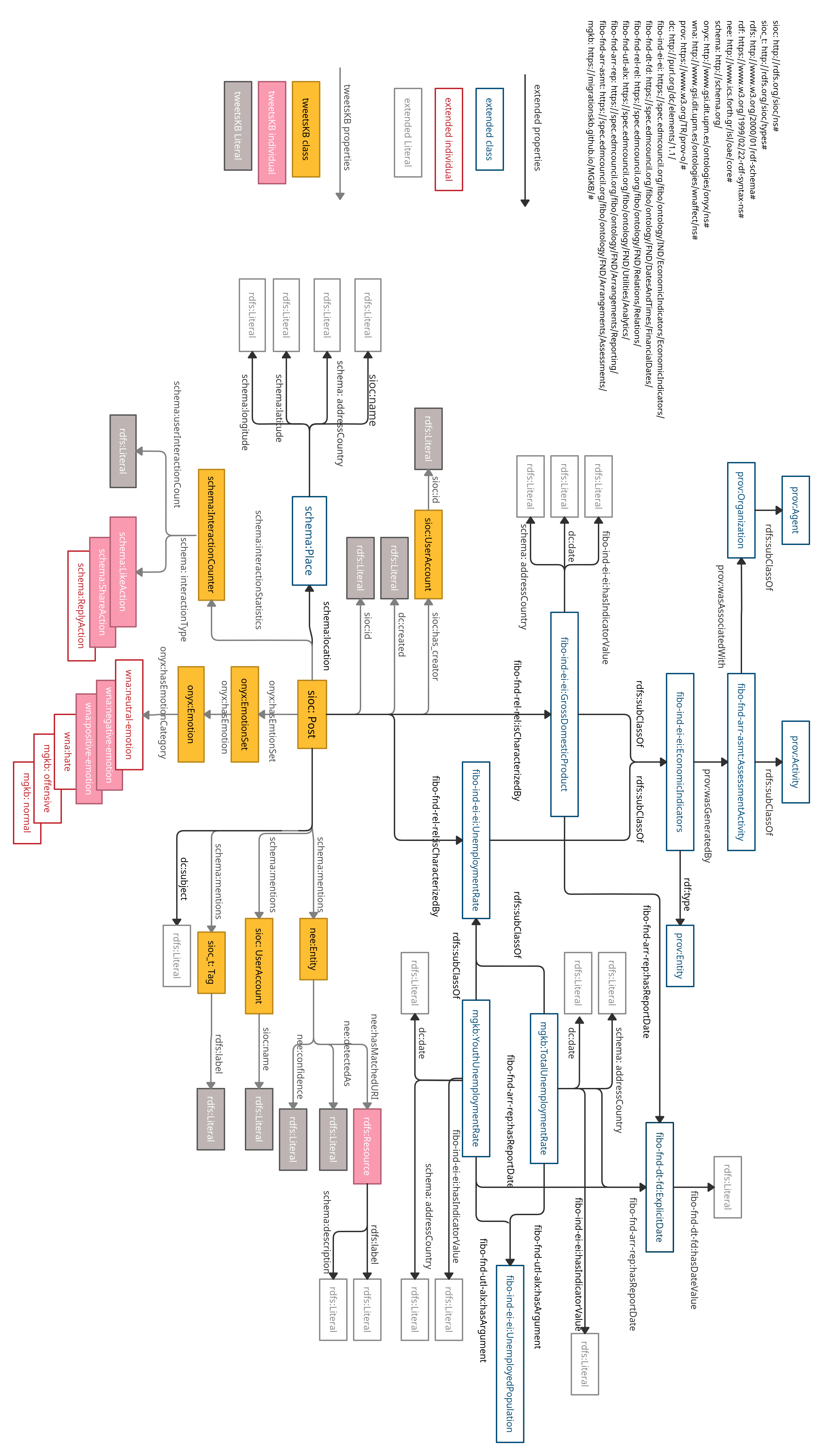}
    \caption{RDF/S Model of {\tt MigrationsKB}.}
    \label{fig:schema}
\end{figure*}

Figure~\ref{fig:schema} shows the RDF/S model of {\tt MigrationsKB}. In order to fulfil the purpose of this study, several classes from already existing ontologies are re-used. A complete documentation on of this RDF/S data model is available online\footnote{\url{https://bit.ly/3oERRSL}}.

For incorporating the meta-data about the geographical location, following information is modified in the TweetsKB. The class {\tt schema:Place} represents Geo information of a tweet. {\tt schema:location} is used for associating a tweet (represented as {\tt sioc:Post}) with {\tt schema:Place}, i.e., its Geo information. {\tt sioc:name} from SIOC\footnote{\url{http://rdfs.org/sioc/spec/}} associates a place with its name represented as a text literal.
{\tt schema:addressCountry} specifies the country code of the geographic location of the tweet. {\tt schema:latitude} and {\tt schema:longitude} specify the coordinates of the Geo information.

\paragraph{Representing Economic Indicators of European Union.}
To represent the economic indicators of the destination countries obtained from Eurostat, Financial Industry Business Ontology (FIBO)\footnote{\url{https://spec.edmcouncil.org/fibo/ontology}} is used. 
        \begin{itemize}
            
            \item The class {\tt fibo-ind-ei-ei:GrossDomesticProduct} represents the GDPR of the country of the tweet in a certain year, which are specified by the properties {\tt schema:addressCountry} and {\tt dc:date} from DCMI, and the value of this indicator is represented by {\tt fibo-ind-ei-ei:hasIndicatorValue}.
            
            \item  The class {\tt fibo-ind-ei-ei:UnemploymentRate}  represents the unemployment rate in the country of the tweet in a certain year, represented with the help of the same properties, i.e.,  {\tt schema:addressCountry}, {\tt dc:date}, and {\tt fibo-ind-ei-ei:hasIndicatorValue}.

            \item The class {\tt fibo-ind-ei-ei:UnemployedPopulation} is used to specify the population of the unemployment rate. 

            \item The class {\tt fibo-fnd-dt-fd:ExplicitDate} represents the date when the statistics are last updated as a literal with the help of the property {\tt fibo-fnd-dt-fd:hasDateValue}.

            \item The property {\tt fibo-fnd-rel-rel:isCharacterizedBy}  is used to associate a tweet with the economic indicators.
        \end{itemize}

\paragraph{Representing Provenance Information.}
To represent the provenance information about the economic indicators, i.e., Eurostat, Statista, UK parliament, and Office of National Statistics, PROV-O\footnote{\url{https://www.w3.org/TR/prov-o/}} is used. The class {\tt prov:Activity} defines an activity that occurs over a period of time and acts upon entities, which are defined by the class {\tt prov:Entity}. The class {\tt fibo-fnd-arr-asmt:AssessmentActivity} represents an assessment activity involving the evaluation and estimation of the economic indicators, which is a subclass of the class {\tt prov:Activity}.
The class {\tt prov:Organization} represents a governmental organization or a company that is associated with the assessment activity, which is a subclass of the class {\tt prov:Agent}. Further extensions are as follows:

\begin{itemize}
    \item {\tt dc:subject} represents a topic of a tweet resulting from topic modeling (see Section~\ref{sec:topic_modeling}).
    \item {\tt wna:neutral-emotion} represents the neutral sentiment of the tweet by applying sentiment analysis.
    \item {\tt wna:hate}, {\tt mgkb:offensive} and {\tt mgkb:normal} represent the hate speeches, offensive speeches and normal speeches from hate speech detection of the tweets. 
    \item {\tt schema:ReplyAction} represents the action of reply regarding a tweet.
    \item {\tt mgkb:EconomicIndicators} represents the economic indicators, which has the subclasses {\tt fibo-ind-ei-ei:GrossDomesticProduct} and {\tt fibo-ind-ei-ei:UnemploymentRate}.
    \item {\tt mgkb:YouthUnemploymentRate} and 
    {\tt mgkb:TotalUnemploymentRate} represent the unemployment rates with respect to the population, i.e., the youth unemployment population and the total unemployed population.
\end{itemize}



\subsection{Competency Questions.}

{\tt MigrationsKB} can further be used for answering the competency questions with the help of the SPARQL Queries, which can be used for querying information from {\tt MigrationsKB}. The following query retrieves the top 10 hashtags which contain ``refugee" and ``immigrant". The query result is shown in Table~\ref{tab:hashtag_refugee}.

\begin{verbatim}
SELECT ?hashtagLabel (count(distinct ?tweet) as ?num) WHERE {
    ?tweet schema:mentions ?hashtag.
    ?hashtag a sioc_t:Tag ; rdfs:label ?hashtagLabel.  
    FILTER( regex(?hashtagLabel, "refugee", "i") || lcase(str(?hashtagLabel))="refugee" 
    ||  regex(?hashtagLabel, "immigrant", "i") || lcase(str(?hashtagLabel))="immigrant").
} GROUP BY ?hashtagLabel ORDER BY DESC(?num) LIMIT 10
\end{verbatim}

\begin{table}[htb]
    \centering
        \caption{The Query Result of retrieving the top 10 hashtags which contain ``refugee" and ``immigrant".  }

    \begin{tabular}{|c|c|}
    \hline 
      
      \textbf{Hashtag}  & \textbf{Nr. of Tweets}  \\\hline 
        refugees & 1354 \\\hline 
        RefugeesWelcome & 1066 \\\hline 
        Refugees & 638 \\\hline 
        refugee & 493 \\\hline 
        RefugeeForum & 372 \\\hline 
        refugeeswelcome &  356  \\\hline 
        WorldRefugeeDay & 333 \\\hline 
        immigrants & 210 \\\hline 
        RefugeeWeek & 183 \\\hline 
        WithRefugees & 148 \\\hline 
        
    \end{tabular}
    \label{tab:hashtag_refugee}
\end{table}

The following query retrieves the top 10 detected entities containing ``refugee". The query result is shown in Table~\ref{tab:entity_refugee}.

\begin{verbatim}
SELECT ?entityLabel (count(?entityLabel) as ?numOfEntityMentions)   where{
    ?tweet schema:mentions ?entity.
	?entity a nee:Entity; nee:hasMatchedURI ?uri. 
    ?uri a rdfs:Resource; rdfs:label ?entityLabel. 
    FILTER( regex(?entityLabel, "refugee", "i") || lcase(str(?entityLabel))="refugee").
 }GROUP BY ?entityLabel ORDER BY DESC(?numOfEntityMentions) LIMIT 10

\end{verbatim}

\begin{table}[htb]
    \centering
        \caption{The Query Result of retrieving the top 10 entities which contain ``refugee" and ``immigrant". }

    \begin{tabular}{|c|c|}
    \hline 
    \textbf{Entity} & \textbf{Nr. of Tweets} \\\hline
      United Nations High Commissioner for Refugees   &  791 \\\hline 
      Refugee   & 183 \\\hline 
      Refugee Nation & 149 \\\hline 
      Refugee Week & 120 \\\hline 
      Convention Relating to the Status of Refugees & 115 \\\hline 
      World Refugee Day & 105 \\\hline 
      Refugee camp & 46 \\\hline 
      European Refugee Fund & 35 \\\hline 
      Refugee Council & 34 \\\hline 
      Refugee Studies Centre & 33 \\\hline 
    \end{tabular}
    \label{tab:entity_refugee}
\end{table}

The following query identifies the sentiments and hatred of the people concerning refugees, i.e. search entities defining ``refugee". The query result is shown in Table~\ref{tab:query_refugee}.


\begin{verbatim}
SELECT ?EmotionCategory (count(distinct ?tweet) as ?numOfTweets) WHERE{
    ?tweet schema:mentions ?entity. 
    ?entity a nee:Entity; nee:hasMatchedURI ?uri. 
    ?uri a rdfs:Resource; rdfs:label ?x. 
        FILTER( regex(?x, "refugee", "i") || lcase(str(?x))="refugee").
  	 ?tweet onyx:hasEmotionSet ?y.
  	 ?y a onyx:EmotionSet; onyx:hasEmotion ?z.
  	 ?z a onyx:Emotion; onyx:hasEmotionCategory ?EmotionCategory.
 } GROUP BY ?EmotionCategory
\end{verbatim}

\begin{table}[htb]
    \centering
        \caption{The Query Result of identifying the Sentiments and Hate Speech Emotions of the Public by searching entities defining ``Refugees".  }

    \begin{tabular}{|c|c|}
    \hline 
      
      \textbf{Emotion Category}  & \textbf{Nr. of Tweets}  \\\hline 
        {\tt wna:neutral-emotion} & 1062 \\\hline 
        {\tt wna:posiive-emotion} & 714 \\\hline 
        {\tt wna:negative-emotion} & 253 \\\hline 
        {\tt mgkb:normal} & 1984 \\\hline 
        {\tt mgkb:offensive} & 8 \\\hline 
        {\tt wna:hate} &  37 \\\hline 
    \end{tabular}
    \label{tab:query_refugee}
\end{table}

The following query retrieves the GDPR indicator values and the number of tweets identified as hate speeches in the United Kingdom by year. The least amount of tweets occurs in 2017 and 2018 (shown in Figure~\ref{fig:dist_after_etm}), there is the least amount of the tweets classified in the ``hate" class during hate speech detection. In the years 2019 and 2020, with the sharp decrease in the GDPR, there are many more hateful tweets than the previous years on yearly basis.
 The query result is shown in Table~\ref{tab:uk_gdpr_hate}. 
 
    
\begin{verbatim}
SELECT  ?year ?IndValue (count(?tweet) as ?numOfTweets) where {
    ?tweet fibo_fnd_rel_rel:isCharacterizedBy ?gdpr.
    
    ?gdpr a fibo_ind_ei_ei:GrossDomesticProduct.
    ?gdpr schema:addressCountry "GB". 
    ?gdpr dc:date ?year.
    ?gdpr fibo_ind_ei_ei:hasIndicatorValue ?IndValue.
    ?tweet onyx:hasEmotionSet ?y.
    ?y a onyx:EmotionSet; onyx:hasEmotion ?z.
    ?z a onyx:Emotion; onyx:hasEmotionCategory wna:hate.
 }GROUP BY ?year ?IndValue ORDER BY DESC(?year)
\end{verbatim}

\begin{table}[htb]
    \centering
    \caption{The GDPR and the Number of Hate Speeches in the United Kingdom. }

    \begin{tabular}{|c|c|c|}
    \hline 
     \textbf{Year}   & \textbf{GDPR} &  \textbf{Nr. of Tweet} \\\hline
     
        2020 &	-9.7 &	5928  \\\hline
        2019 &	1.4  &	4698  \\\hline
        2018 &	1.3	  & 126  \\\hline
        2017 &	1.7	  & 115   \\\hline
        2016  &	1.7	 & 299    \\\hline
        2015 &	2.4  &	951    \\\hline
        2014 &	2.9	 & 1865   \\\hline
        2013  &	2.2  &	934 \\\hline
        \end{tabular}
    \label{tab:uk_gdpr_hate}
\end{table}

\section{Discussion and Future Work}
\label{sec:discussion}

In the current study, a Knowledge Base of migration related tweets is represented. The tweets are filtered and annotated using BERT-based sentiment analysis and attention-based hate speech detection. {\tt MigrationsKB} extends the RDF/S model defined by TweetsKB by adding the Geo information of tweets, the statistics of economic indicators of European Union, and the results from hate speech detection algorithm. The corpus would assist further research in various fields such as social science by providing readily available information. 
With the integrated features, the relations between the public attitudes towards migrations and the economic factors have been made query-able. 

As for now, the focus is solely on the English tweets, the distribution of the corpus is therefore highly skewed by the tweets from the United Kingdom. While focusing on the destination countries in Europe, there is already a wide variety of languages that need attention. 
Secondly, visualization tools and interfaces for querying will be created to help the experts in other fields such as social scientists to effectively interact with {\tt MigrationsKB}. 
Finally, the {\tt MigrationsKB} will be under continuous development by updating with newer relevant tweets. The more advanced approaches for topic modeling will be experimented, to analyze the change of topics in tweets in the temporal dimension. 

\section*{Acknowledgement}

This work is a part of ITFlows project\footnote{\url{https://www.itflows.eu/}}. This project has received funding from the European Union's Horizon 2020 research and innovation programme under grant agreement No 882986.


\end{document}